\documentclass[10pt,twocolumn,letterpaper]{article}

\usepackage[pagenumbers]{cvpr} 

\usepackage{graphicx}
\usepackage{amsmath}
\usepackage{amssymb}
\usepackage{booktabs}
\usepackage{bbm}
\usepackage{multirow}
\usepackage{subcaption}
\usepackage{url}
\usepackage{paralist}

\makeatletter
\def\hlinew#1{%
	\noalign{\ifnum0=`}\fi\hrule \@height #1 \futurelet
	\reserved@a\@xhline}
\makeatother%
\usepackage[pagebackref=true,breaklinks=true,letterpaper=true,colorlinks,citecolor=blue,linkcolor=red,bookmarks=false]{hyperref}

\usepackage[capitalize]{cleveref}
\crefname{section}{Sec.}{Secs.}
\Crefname{section}{Section}{Sections}
\Crefname{table}{Table}{Tables}
\crefname{table}{Tab.}{Tabs.}
\def\cvprPaperID{506} 
\def\confName{CVPR}
\def\confYear{2023}

\begin{document}

\title{Learning A Sparse Transformer Network for Effective Image Deraining}

\author{Xiang Chen$^{1}$\quad Hao Li$^{1}$\quad Mingqiang Li$^{2}$\quad Jinshan Pan$^{1}$\thanks{Corresponding author.} \\
	$^{1}$School of Computer Science and Engineering, Nanjing University of Science and Technology \\
$^{2}$Information Science Academy, China Electronics Technology Group Corporation \\
\\
}

\maketitle

\begin{abstract}
Transformers-based methods have achieved significant performance in image deraining as they can model the non-local information which is vital for high-quality image reconstruction.
In this paper, we find that most existing Transformers usually use all similarities of the tokens from the query-key pairs for the feature aggregation. However, if the tokens from the query are different from those of the key, the self-attention values estimated from these tokens also involve in feature aggregation, which accordingly interferes with the clear image restoration.
To overcome this problem, we propose an effective \textbf{D}e\textbf{R}aining network, \textbf{S}parse Trans\textbf{former} (DRSformer) that can adaptively keep the most useful self-attention values for feature aggregation so that the aggregated features better facilitate high-quality image reconstruction.
Specifically, we develop a learnable top-k selection operator to adaptively retain the most crucial attention scores from the keys for each query for better feature aggregation.
Simultaneously, as the naive feed-forward network in Transformers does not model the multi-scale information that is important for latent clear image restoration, we develop an effective mixed-scale feed-forward network to generate better features for image deraining.
To learn an enriched set of hybrid features, which combines local context from CNN operators, we equip our model with mixture of experts feature compensator to present a cooperation refinement deraining scheme.
Extensive experimental results on the commonly used benchmarks demonstrate that the proposed method achieves favorable performance against state-of-the-art approaches.
The source code and trained models are available at \url{https://github.com/cschenxiang/DRSformer}.

\vspace{-4mm}
\end{abstract}

\vspace{-1mm}
\section{Introduction}
Single image deraining is a typical low-level vision problem emerging in the last decade. It aims to recover the clean image from the observed rainy one.
As the clear image and rain streaks are unknown, it is an ill-posed inverse problem.
To solve this problem, early approaches \cite{kang2011automatic,li2016rain,zhang2017convolutional} usually impose various priors based on statistical properties of rain streaks and clear images.
In fact, these handcrafted priors are not robust to complex and varying rainy scenarios, which limit the deraining performance.

\begin{figure}[!t]
	\centering 	
	\begin{subfigure}[t]{0.32\columnwidth}
		\centering
		\includegraphics[width=\columnwidth]{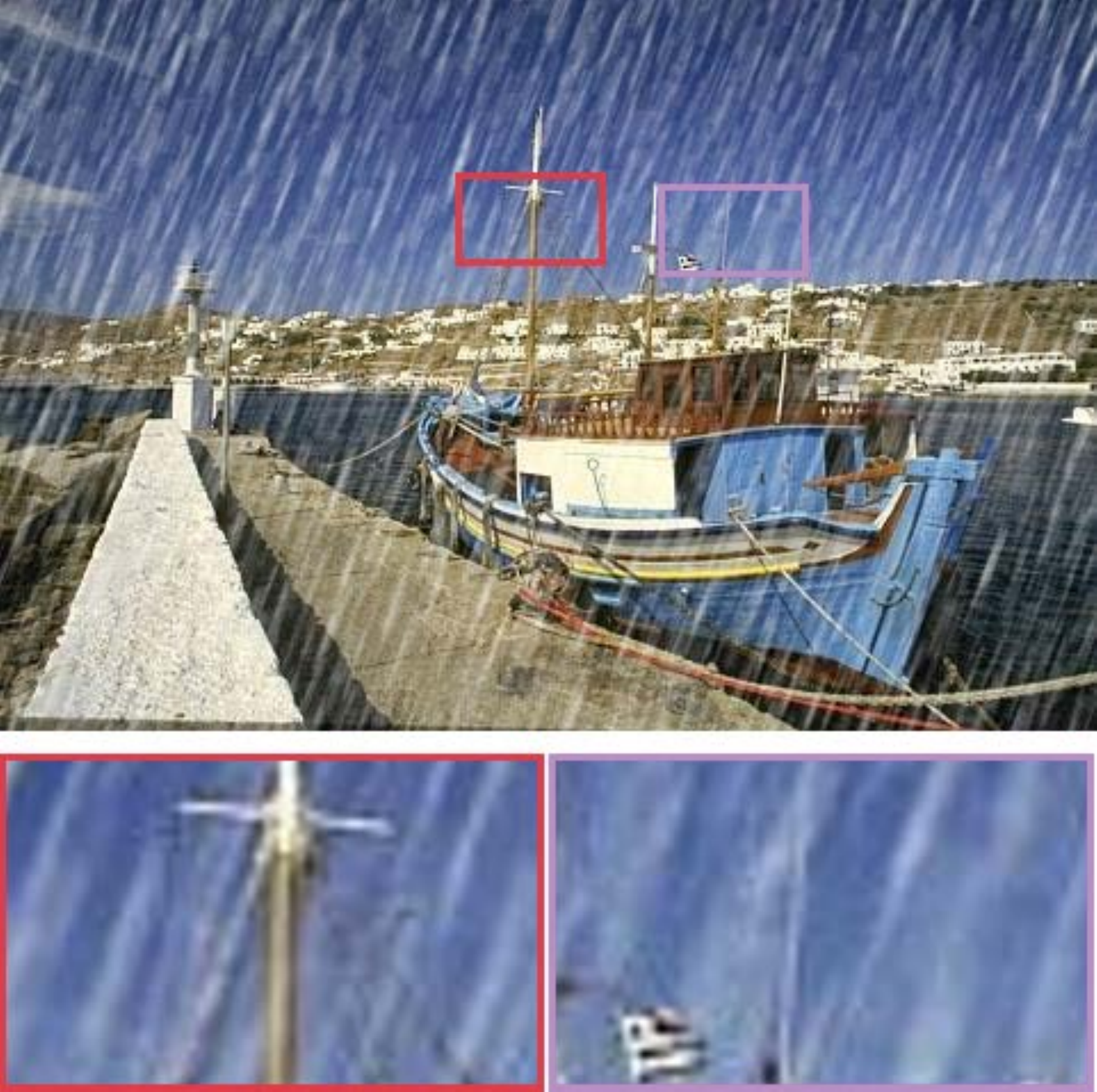}
		\caption{Rainy Input}
	\end{subfigure}
	\begin{subfigure}[t]{0.32\columnwidth}
		\centering
		\includegraphics[width=\columnwidth]{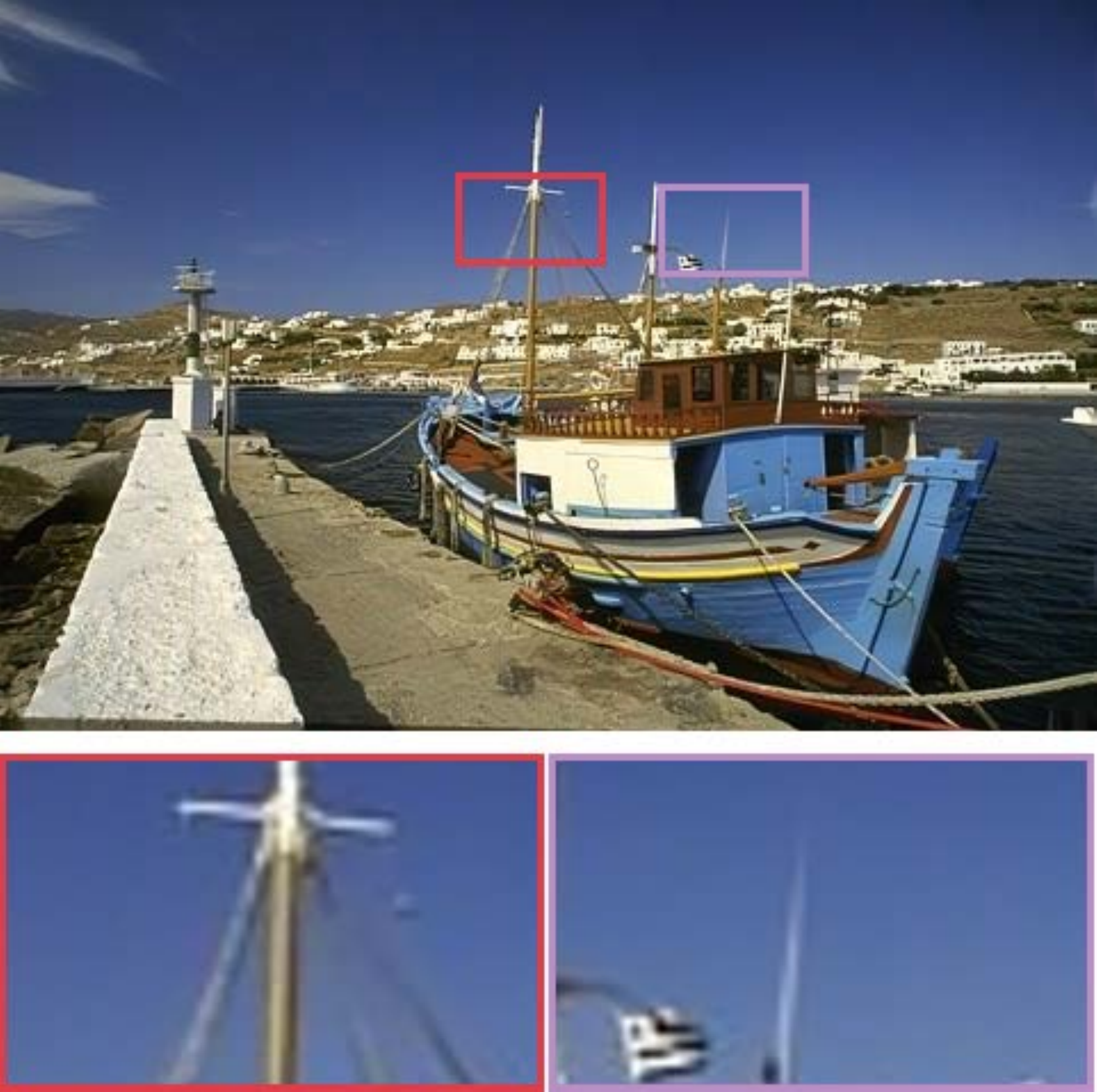}
		\caption{Uformer~\cite{wang2022uformer}}
	\end{subfigure}
	\begin{subfigure}[t]{0.32\columnwidth}
	\centering
	\includegraphics[width=\columnwidth]{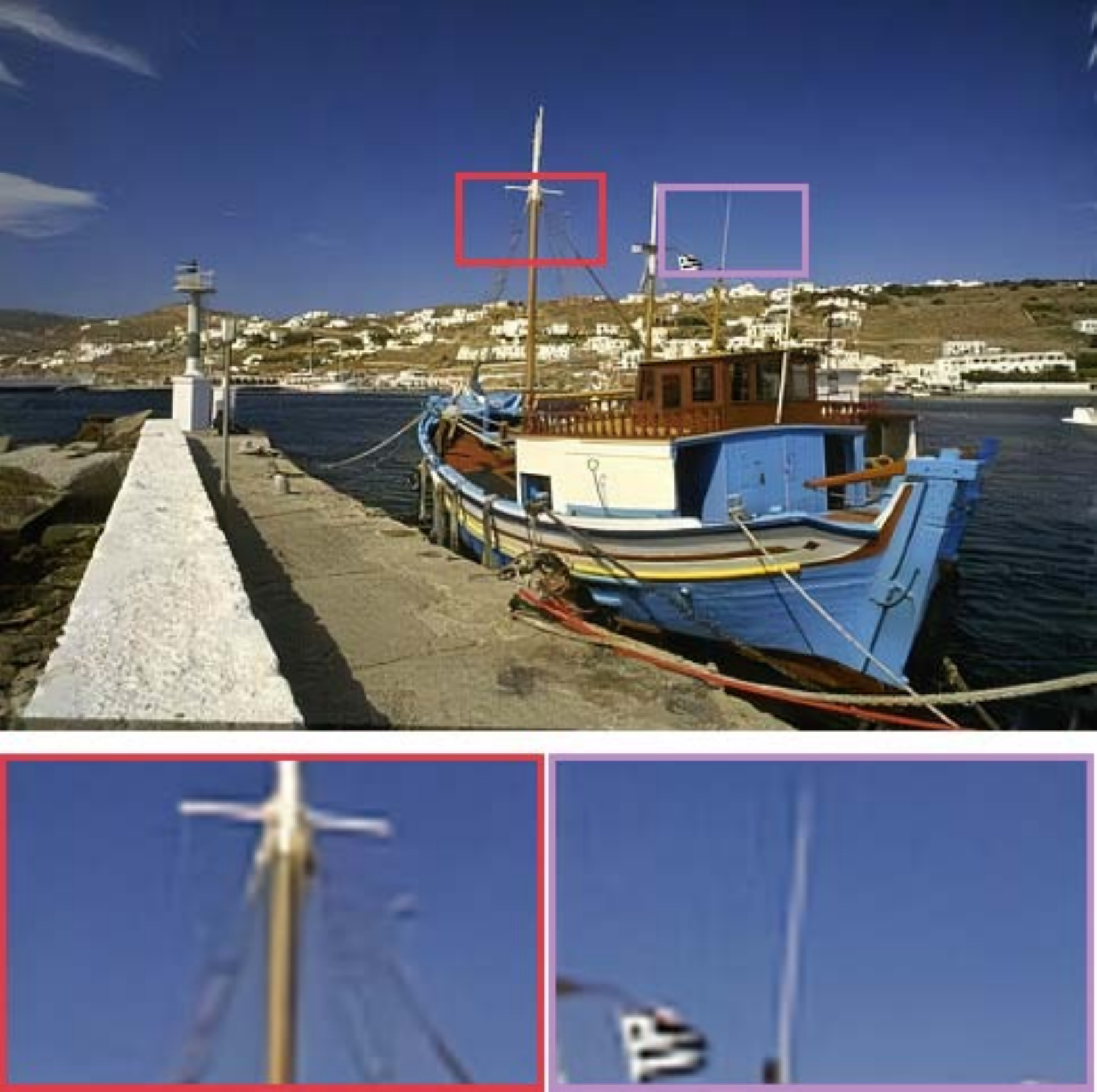}
	\caption{Restormer~\cite{zamir2022restormer}}
    \end{subfigure}	
    \\
    \begin{subfigure}[t]{0.32\columnwidth}
    \centering
    \includegraphics[width=\columnwidth]{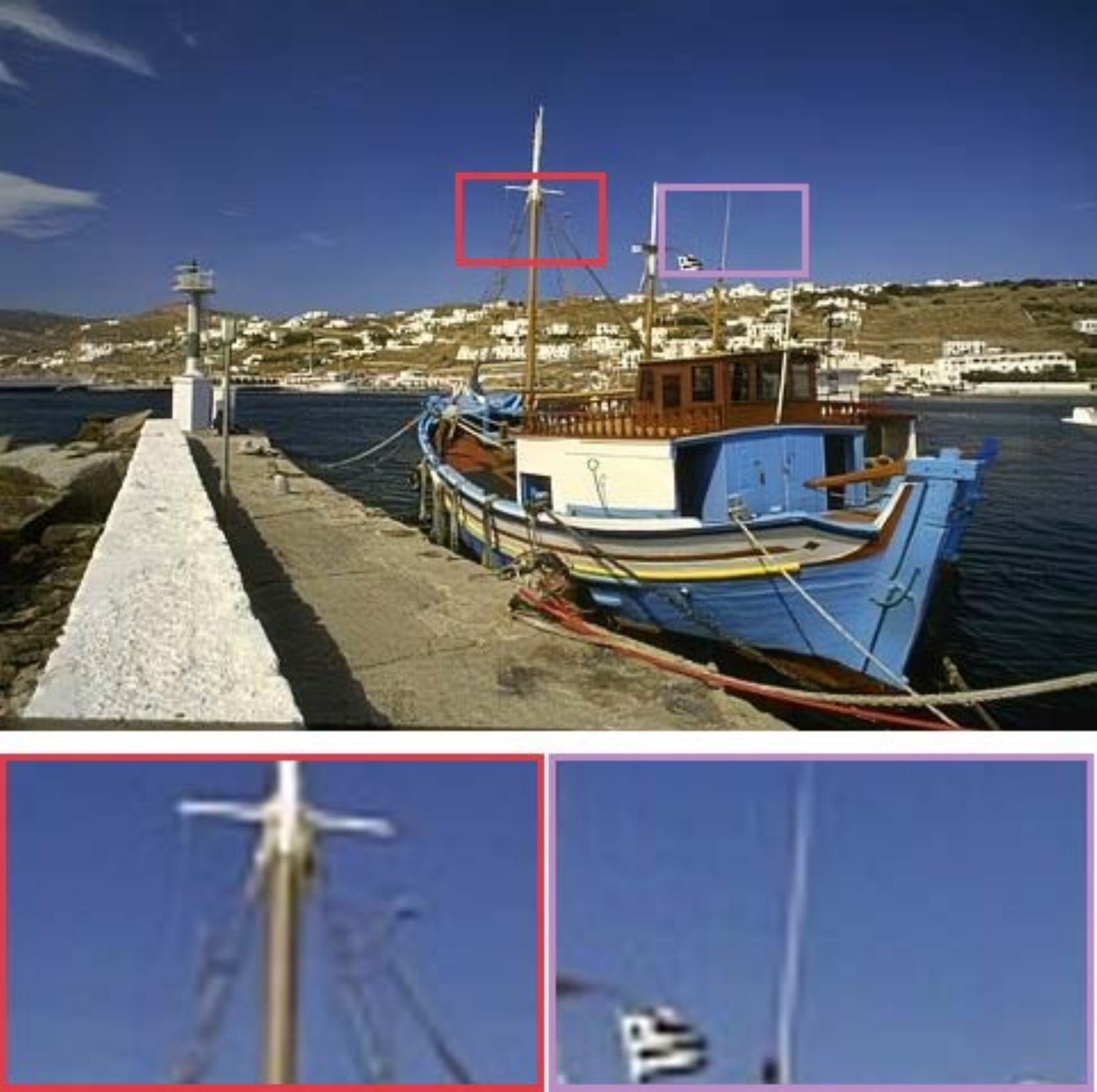}
    \caption{IDT~\cite{xiao2022image}}
    \end{subfigure}
    \begin{subfigure}[t]{0.32\columnwidth}
    \centering
    \includegraphics[width=\columnwidth]{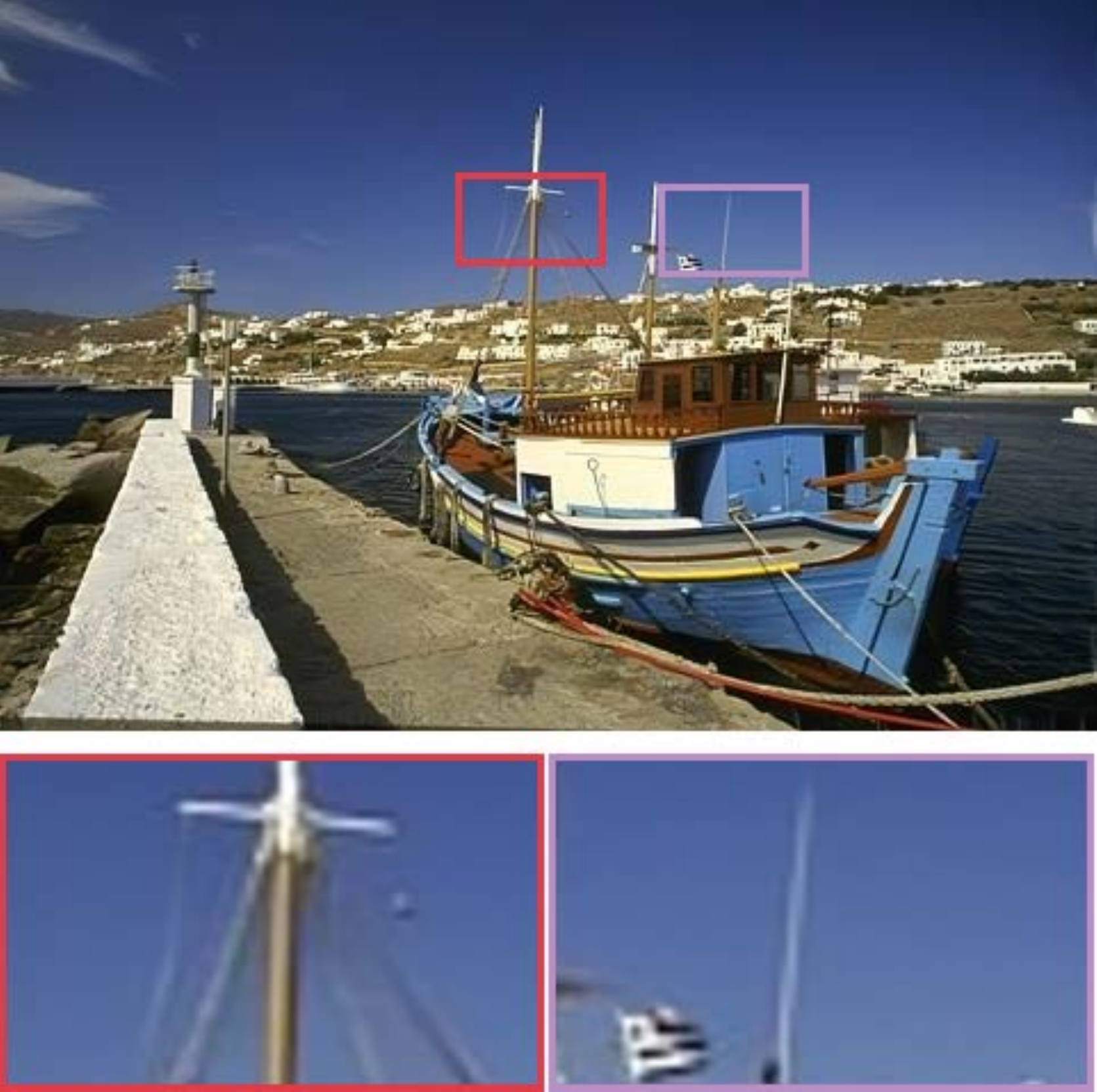}
    \caption{Ours}
    \end{subfigure}
    \begin{subfigure}[t]{0.32\columnwidth}
    \centering
    \includegraphics[width=\columnwidth]{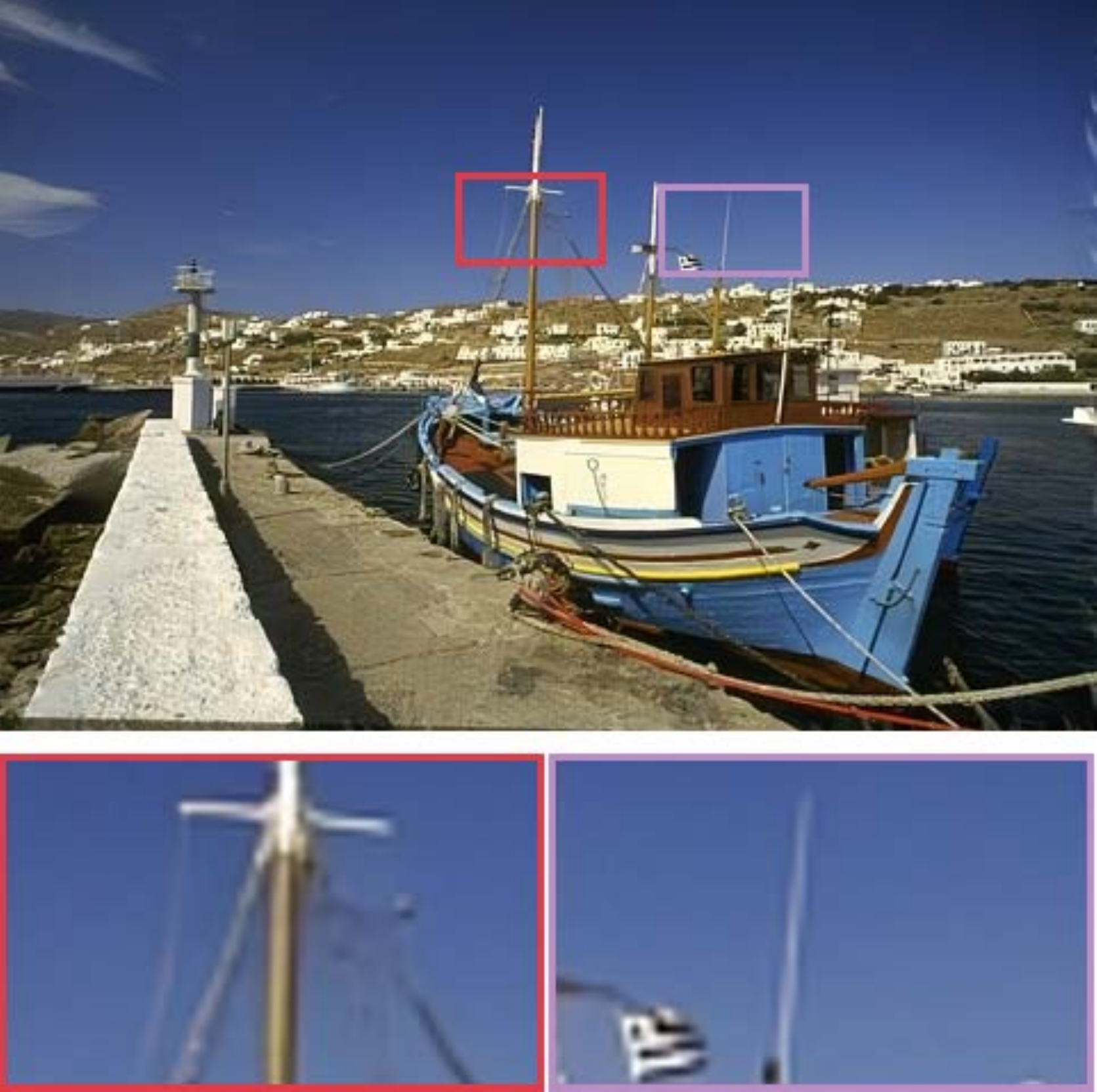}
    \caption{Ground Truth}
    \end{subfigure}
	\vspace{-2mm}
	\caption{Image deraining results between our method and recent Transformer-based methods \cite{wang2022uformer,zamir2022restormer,xiao2022image}. Our method can generate high-quality image with more accurate detail and texture recovery.}
	\label{fig1}
	\vspace{-4mm}
\end{figure}

Recently, numerous learning-based methods \cite{yang2017deep, li2018recurrent, ren2019progressive, jiang2020multi, yang2020single, yi2021structure, chen2022unpaired} have resorted to diverse CNN architectures as a preferable choice compared to traditional algorithms.
However, the intrinsic characteristics of convolutional operation, \emph{i.e}, local receptive fields and independence of input content, hinder the model's capacity to eliminate long-range rain degradation perturbation.
To alleviate such limitations, Transformers \cite{chen2021pre, xiao2022image, qin2021etdnet, liang2022drt} have been applied to image deraining and have achieved decent performance as they can better model the non-local information for high-quality image reconstruction.
Nevertheless, the image details, which are local features of images, are not modeled well by these approaches when restoring clear images as shown in Figure~\ref{fig1}.
One main reason is that the self-attention in Transformers does not model the local invariant properties that CNNs do well.
Since rain streaks tend to confuse with background details in local regions, recent studies \cite{yuan2021incorporating, jiang2022magic, chen2023hybrid} try to mitigate such drawbacks by combining CNN operations and Transformers for boosting image deraining, where the Transformers based on the standard formulations.

We note that the standard Transformers \cite{vaswani2017attention} usually use all attention relations based on the query-key pairs to aggregate features.
As the tokens from the key are not always relevant to those from the query, using the self-attention values estimated from these tokens in the feature aggregation interferes with the following latent clear image restoration.
The root cause behind this deficiency lies in that, the native dense calculation pattern of self-attention amplifies relatively smaller similarity weights, making feature interaction and aggregation process susceptible to implicit noises.
This also naturally leads to corresponding redundant or irrelevant representations are still taken into consideration when modeling global feature dependencies \cite{zhao2019explicit, wang2022kvt}.
Thus, these findings motivate us to explore the most useful self-attention values so that we can make full use of the features for better image restoration.

To this end, we develop an effective sparse Transformer network for image deraining, named as DRSformer.
Specifically, the key component of the proposed framework is the sparse Transformer block (STB) which contains a top-\emph{k} sparse attention (TKSA) that keeps the most useful self-attention values for feature aggregation and a mixed-scale feed-forward network (MSFN) that explores the multi-scale features for better image deraining.
First, we design the top-\emph{k} attention mechanism to replace the vanilla self-attention \cite{vaswani2017attention}.
The TKSA keeps the largest $K$ similarity scores between the queries and the keys for the self-attention computing, thereby facilitating better feature aggregation.
Furthermore, the developed MSFN further explores the multi-scale information to better improve the aggregated features.
Finally, based on the observation that rain distribution reveals the degradation location and degree, we also introduce mixture of experts feature compensator (MEFC) to provide collaborative refinement for STB.
With the above-mentioned designs, our proposed method offers three-fold advantages: (1) it can enjoy natural robustness in terms of less sensitivity to useless feature interference, (2) it can not only enrich the locality but also empower the capability of global feature exploitation, and (3) it can co-explore data (embodied in MEFC) and content (embodied in STB) sparsity for achieving deraining performance gains.

The main contributions are summarized as follows:
\begin{compactitem}
	\item We propose a sparse Transformer architecture to help generate high-quality deraining results with more accurate detail and texture recovery.

	\item We develop a simple yet effective learnable top-\emph{k} selection operator to adaptively maintain the most useful self-attention values for better feature aggregation.

	\item We design an effective feed-forward network based on mixed-scale fusion strategy to explore multi-scale representations for better facilitating image deraining.

	\item Extensive experimental results on various benchmarks demonstrate that our method achieves favorable performance against state-of-the-art (SOTA) approaches.
\end{compactitem}

\section{Related Work}
\vspace{-1mm}
{\flushleft\textbf{Single image deraining}.}
Since image deraining is an ill-posed problem, traditional methods~\cite{kang2011automatic,li2016rain,zhang2017convolutional,luo2015removing, gu2017joint} usually develop kinds of image priors to provide additional constraints.
However, these handcrafted priors tend to rely on empirical observations and thus are not able to model the inherent properties of clear images.
To overcome this problem, numerous CNN-based frameworks \cite{yang2020single} have been developed to solve image deraining and achieved decent restoration performance.
To better represent the rain distribution, several studies take rain characteristics such as rain direction \cite{liu2021unpaired}, density \cite{zhang2018density}, veiling effect \cite{hu2019depth} into account, and optimize the network structure via recursive computation \cite{li2018recurrent, ren2019progressive, jiang2020multi} or transfer mechanism \cite{wei2019semi, yasarla2020syn2real, ye2021closing, huang2021memory}.
Although these methods achieve better performance than the hand-crafted prior-based ones, they have difficulty capturing the long-range dependencies due to the intrinsic limitations of convolution.
Different CNN-based deraining approaches, we utilize the Transformer as the network backbone to model non-local information for image deraining.

\vspace{-2mm}
{\flushleft\textbf{Vision Transformers}.}
Motivated by the great success of the Transformers \cite{dosovitskiy2020image} in natural language processing (NLP) \cite{vaswani2017attention} and high-level vision tasks \cite{carion2020end, liu2021swin}, Transformers have been applied to image restoration \cite{2020learning, chen2021pre, zamir2022restormer, wang2022uformer, guo2022image} and perform better than the previous CNN-based baselines as they are able to model non-local information.
For the field of image rain removal, Jiang {\em et al.} \cite{jiang2022magic} design a dynamic associated deraining network by incorporating self-attention in Transformer with a background recovery network. More recently, Xiao {\em et al.} \cite{xiao2022image} elaborately develop image deraining Transformer (IDT) with window-based and spatial-based dual Transformer to achieve excellent results.
Note that, most existing methods rely on the dense dot-product self-attention as the heart of Transformers. However, one shortcoming of this computation manner is that redundant or irrelevant features with smaller weights may interfere with the attention map, which makes the output features contain potential noises. In this work, we propose sparse attention in Transformer to relieve the negligence of the most relevant information faced by vanilla self-attention.

\begin{figure*}[t]
	\centering
	\includegraphics[width=1.0\textwidth]{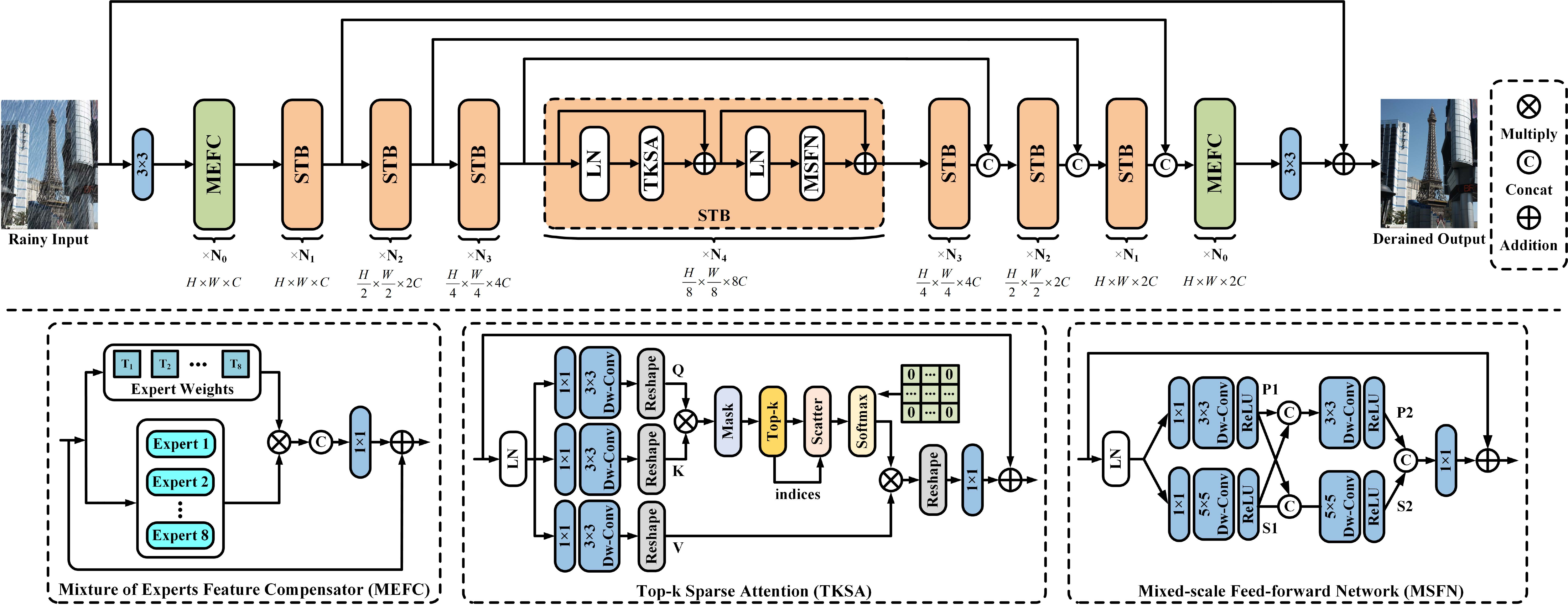}
	\vspace{-5mm}
	\caption{The overall architecture of the proposed sparse Transformer network for image deraining (DRSformer), which mainly contains sparse Transformer block (STB) with top-\emph{k} sparse attention (TKSA) and mixed-scale feed-forward network (MSFN), and mixture of experts feature compensator (MEFC). LN refers to the layer normalization and DW-Conv refers to the depth-wise convolution.}
	\label{fig2}
	\vspace{-5mm}
\end{figure*}

\vspace{-2mm}
{\flushleft\textbf{Sparse representation}.}
With inspirations drawn from neural activity in biological brains, sparsity of hidden representation in deep neural networks as a tantalizing ``free lunch'' emerges for both vision and NLP tasks \cite{zhao2019explicit, wang2022kvt}.
Indeed, it is widely proven that sparse representation also plays a critical role in handling low-level vision problems, such as image deraining \cite{wang2021multi} and super-resolution \cite{mei2021image}.
In principle, sparse attention can be categorized into data-based (fixed) sparse attention and content-based sparse attention \cite{correia2019adaptively, roy2021efficient}.
For data-based sparse attention, several local attention operations are introduced into CNN backbone, which mainly considers attending only to local window size.
Recent studies \cite{wang2022nformer, fu2022sparsett} have investigated enforcing sparsity to Transformer backbone.
More recently, Zhang {\em et al.} \cite{zhang2023accurate} design an attention retractable Transformer to allow tokens from sparse areas to interact features, which is data-based sparsity.
Different from it, we implement a simple but effective approximation for self-attention based on top-\emph{k} selection to achieve sparse attention, which is content-based sparsity.

\vspace{-2mm}
{\flushleft\textbf{Top-k selection}.}
Zhao {\em et al.} \cite{zhao2019explicit} first propose an explicit selection method based on top-\emph{k} mechanism in NLP tasks.
Driven by their success, \emph{k}-NN attention \cite{wang2022kvt,wang2022nformer} is further introduced for boosting vision Transformers.
Unlike performing top-\emph{k} selection in the spatial dimension \cite{wang2022kvt}, we design an efficient top-\emph{k} useful channel selection operator.

\vspace{-1mm}
\section{Proposed Method}
\vspace{-1mm}
In this section, we first describe the overall pipeline and symmetrically hierarchical network architecture for image deraining.
Afterward, we provide the details of the proposed sparse Transformer block (STB), as the fundamental building unit of our method, which mainly contains two key elements: top-\emph{k} sparse attention (TKSA) and mixed-scale feed-forward network (MSFN).
Finally, we present the introduced mixture of experts feature compensator (MEFC).
%

\subsection{Overall pipeline}
The overall pipeline of our proposed DRSformer, shown in Figure~\ref{fig2}, is based on a hierarchical encoder-decoder framework.
Given a rainy image $I_{rain} \in \mathbb{R}^{H \times W \times 3}$, where $H \times W$ represents the spatial resolution of the feature map, we perform overlapped image patch embedding with $3 \times 3$ convolution.
In the network backbone, we stack $N_{i \in[1,2,3,4]}$ STBs to extract rich features for spatially-varying rain distribution.
To excavate the multi-scale representation from rain degeneration, each level of encoder-decoder pipeline covers its own specific spatial resolution and channel dimension.
For feature down-sampling and up-sampling, we apply pixel-unshuffle and pixel-shuffle operations.
Similar to \cite{zamir2022restormer, wang2022uformer, xiao2022image}, we also add skip-connections to bridge across continuous intermediate features for stable training.
In each STB, unlike the standard self-attention \cite{dosovitskiy2020image} in Transformer, we develop TKSA to achieve feature sparsity, aiming to enforce the feature aggregation process more effectively.
In addition, a MSFN is introduced into STB to enrich multi-scale local information and help image restoration.
At the early and final stages of the model learning, we equip our model with $N_{0}$ MEFCs to provide complementary feature refinement, so that high-quality clear outputs can be finally reconstructed.
With this hybrid formulation, we allow DRSformer to exploit both the adaptive content and the intrinsic property of rainy images, facilitating the separation of undesired rain streaks and latent clear background, and experiments demonstrate that the above design choices yield quality improvements (see Sec. \ref{4.3})

The final reconstructed result is obtained by: $I_{derain} = \mathcal{F}(I_{rain}) + I_{rain}$, where $\mathcal{F}(\cdot)$ is the overall network and it is trained by minimizing the following loss function:
\begin{equation}
	\mathcal{L}=\left\|I_{derain}-I_{gt}\right\|_{1},
\end{equation}
where $I_{gt}$ denotes the ground-truth image, and $\|\cdot\|_{1}$ denotes the $L_1$-norm.

\subsection{Sparse Transformer block}
As the standard Transformers \cite{vaswani2017attention, dosovitskiy2020image, zamir2022restormer} take all the tokens to compute self-attention globally, which is unfriendly for image restoration due to it may involve noisy interactions between the irrelevant features.
To solve such limitations, we develop a sparse Transformer block (STB) as the feature extraction unit by taking the advantages of sparsity that emerged in neural networks \cite{zhao2019explicit}.
Formally, given the input features at the ($l$-1)-th block $\mathbf{X}_{l-1}$, the encoding procedures of STB can be defined as:
\vspace{-2mm}
\begin{equation}
	\mathbf{X}_l^{\prime}=\mathbf{X}_{l-1}+\text{TKSA}\left(\text{LN}\left(\mathbf{X}_{l-1}\right)\right),
\end{equation}
\begin{equation}
	\mathbf{X}_l=\mathbf{X}_l^{\prime}+\text{MSFN}\left(\text{LN}\left(\mathbf{X}_l^{\prime}\right)\right),
\vspace{-1mm}
\end{equation}
where $\text{LN}$ denotes the layer normalization; $\mathbf{X}_l^{\prime}$ and $\mathbf{X}_l$ denote the outputs from the top-\emph{k} sparse attention (TKSA) and mixed-scale feed-forward network (MSFN), which are described below.

\vspace{-2mm}
{\flushleft\textbf{Top-k sparse attention (TKSA)}.}
We revisit the standard self-attention in Transformer, which has become an empirical operation in most of the existing models.
Given a query $Q$, key $K$ and value $V$ with the dimension of $\mathbb{R}^{L \times d}$, the output of dot-product attention is generally formulated as:
\vspace{-1mm}
\begin{equation}
	\text{Att}(\mathbf{Q}, \mathbf{K}, \mathbf{V})=\operatorname{softmax}\left(\frac{\mathbf{Q} \mathbf{K}^{\top}}{\lambda}\right) \mathbf{V},
	\label{4}
\vspace{-1mm}
\end{equation}
where $\mathbf{Q}$, $\mathbf{K}$, and $\mathbf{V}$ denote the matrix forms of $Q$, $K$, and $V$, respectively. $\lambda$ is an optional temperature factor defined by $\lambda = \sqrt{d}$.
Generally, multi-head attention is implemented to each of the $k$ new $Q$, $K$ and $V$, yielding $d=C / k$ channel dimensional outputs which are concatenated and then got the final result for all heads via the linear projection.
Noted that, this vanilla self-attention paradigm is based on densely fully-connected, which requires computing the attention map for all query-key pairs.
In our work, we develop TKSA to replace it, thus avoiding the involvement of irrelevant information during the feature interaction process.

Specifically, we first encode channel-wise context by applying $1 \times 1$ convolutions followed by $3 \times 3$ depth-wise convolutions.
Inspired by \cite{zamir2022restormer}, we apply self-attention across channels rather than the spatial dimension to reduce the time and memory complexity.
Next, we calculate similarities of pixel pairs between all the reshaped queries and keys, and mask out the unnecessary elements assigned with lower attention weights in the transposed attention matrix $\boldsymbol{M}$ of size $\mathbb{R}^{\hat{C} \times \hat{C}}$.
Unlike the dropout strategy of randomly abandoning the scores, an adaptive selection of the top-\emph{k} contributive scores is implemented upon $\boldsymbol{M}$, aiming to preserve the most significant components and remove the useless ones \cite{chen2021chasing}.
Here, \emph{k} is an adjustable parameter to dynamically control the magnitude of sparsity, which is formally obtained by weighted average of some proper fractions, such as $\frac{2}{3}$.
Thus, only top-\emph{k} values within the range $[\Delta_{1}, \Delta_{2}]$ are normalized from each row of $\boldsymbol{M}$ for softmax computing.
For other elements that are smaller than top-\emph{k} scores, we replace their probabilities with 0 at given indices using the scatter function.
This dynamic selection makes the attention from \emph{dense} to \emph{sparse}, which is derived by:
\vspace{-1mm}
\begin{equation}
	\text{SparseAtt}(\mathbf{Q}, \mathbf{K}, \mathbf{V})=\operatorname{softmax}\left(\mathcal{T}_k(\frac{\mathbf{Q} \mathbf{K}^{\top}}{\lambda})\right) \mathbf{V},
\vspace{-1mm}
\end{equation}
where $\mathcal{T}_k(\cdot)$ is the learnable top-\emph{k} selection operator:
\vspace{-1mm}
\begin{equation}
	\left[\mathcal{T}_k(\boldsymbol{S})\right]_{i j}= \begin{cases}S_{i j} & S_{i j} \in \text { top-k(row } j)
	\\ 0 & \text { otherwise. }\end{cases}
\vspace{-1mm}
\end{equation}

Finally, we multiply the $\operatorname{softmax}$ and value by matrix multiplication.
As we use the multi-head strategy, we concatenate all the outputs of multi-head attention, and then get the final result by the linear projection.

{\flushleft\textbf{Mixed-scale feed-forward network (MSFN)}.}
Previous studies \cite{zamir2022restormer, wang2022uformer, xiao2022image} usually introduce single-scale depth-wise convolutions into the regular feed-forward network to imporve locality.
However, those exploitations all ignore the correlations of multi-scale rain streaks.
In fact, rich multi-scale representation has fully demonstrated its effectiveness \cite{jiang2020multi, wang2020dcsfn} in better removing rain.
Here, we design a MSFN by inserting two multi-scale depth-wise convolution paths in the transmission process, see Figure \ref{fig2}.
Given an input tensor $\mathbf{X}_{l-1} \in \mathbb{R}^{H \times W \times C}$, after layer normalization, we first utilize $1 \times 1$ convolution to expand the channel dimension in the ratio of $r$, then feed it into two parallel branches.
During the feature transformation, the $3 \times 3$ and $5 \times 5$ depth-wise convolutions are employed to enhance the multi-scale local information extraction.
In this way, the entire feature fusion procedure of the developed MSFN is formulated as:
\vspace{-1mm}
\begin{equation}
\begin{aligned}
&\hat{\mathbf{X}}_l = f_{1 \times 1}^c\left(\text{LN}\left(\mathbf{X}_{l-1}\right)\right), \\
&\mathbf{X}_{l}^{p_{1}} = \sigma\left(f_{3 \times 3}^{dwc}  (\hat{\mathbf{X}}_l)\right), \mathbf{X}_{l}^{s_{1}} = \sigma\left(f_{5 \times 5}^{dwc} (\hat{\mathbf{X}}_l) \right), \\
&\mathbf{X}_{l}^{p_{2}} = \sigma\left(f_{3 \times 3}^{dwc} [\mathbf{X}_{l}^{p_{1}}, \mathbf{X}_{l}^{s_{1}}] \right), \mathbf{X}_{l}^{s_{2}} = \sigma\left(f_{5 \times 5}^{dwc} [\mathbf{X}_{l}^{s_{1}}, \mathbf{X}_{l}^{p_{1}}] \right), \\
&\mathbf{X}_l = f_{1 \times 1}^c [\mathbf{X}_{l}^{p_{2}}, \mathbf{X}_{l}^{s_{2}}] + \mathbf{X}_{l-1},
\end{aligned}
\vspace{-1mm}
\end{equation}
where $\sigma(\cdot)$ is a ReLU activation, $f_{1 \times 1}^c$ represents $1 \times 1$ convolution, $f_{3 \times 3}^{dwc}$ and $f_{5 \times 5}^{dwc}$ denote $3 \times 3$ and $5 \times 5$ depth-wise convolutions, and $[\cdot]$ is the channel-wise concatenation.

\begin{table*}[t]
	\centering
	\caption{Comparison of quantitative results on synthetic and real datasets. \textbf{Bold} and \underline{underline} indicate the best and second-best results.}
	\vspace{-3mm}
	\resizebox{0.97\textwidth}{!}{
	\begin{tabular}{cc|cc|cc|cc|cc|cc}
	\hlinew{1.0pt}
	\multicolumn{2}{c|}{Datasets}                                                   & \multicolumn{2}{c|}{\textit{Rain200L}}    & \multicolumn{2}{c|}{\textit{Rain200H}}    & \multicolumn{2}{c|}{\textit{DID-Data}}    & \multicolumn{2}{c|}{\textit{DDN-Data}}    & \multicolumn{2}{c}{\textit{SPA-Data}} \\ \hline
	\multicolumn{2}{c|}{Metrics}                                                    & PSNR           & SSIM            & PSNR           & SSIM            & PSNR           & SSIM            & PSNR           & SSIM            & PSNR           & SSIM            \\ \hline
	\multicolumn{1}{c|}{\multirow{2}{*}{Prior-based methods}}       & DSC \cite{luo2015removing}          & 27.16          & 0.8663          & 14.73          & 0.3815          & 24.24          & 0.8279          & 27.31          & 0.8373          & 34.95          & 0.9416          \\
	\multicolumn{1}{c|}{}                                           & GMM \cite{li2016rain}          & 28.66          & 0.8652          & 14.50          & 0.4164          & 25.81          & 0.8344          & 27.55          & 0.8479          & 34.30          & 0.9428          \\ \hline
	\multicolumn{1}{c|}{\multirow{8}{*}{CNN-based methods}}         & DDN \cite{fu2017removing}          & 34.68          & 0.9671          & 26.05          & 0.8056          & 30.97          & 0.9116          & 30.00          & 0.9041          & 36.16          & 0.9457          \\
	\multicolumn{1}{c|}{}                                           & RESCAN \cite{li2018recurrent}       & 36.09          & 0.9697          & 26.75          & 0.8353          & 33.38          & 0.9417          & 31.94          & 0.9345          & 38.11           & 0.9707          \\
	\multicolumn{1}{c|}{}                                           & PReNet \cite{ren2019progressive}       & 37.80          & 0.9814          & 29.04          & 0.8991          & 33.17          & 0.9481          & 32.60          & 0.9459          & 40.16          & 0.9816          \\
	\multicolumn{1}{c|}{}                                           & MSPFN \cite{jiang2020multi}       & 38.58          & 0.9827          & 29.36          & 0.9034          & 33.72          & 0.9550          & 32.99          & 0.9333          & 43.43          & 0.9843          \\
	\multicolumn{1}{c|}{}                                           & RCDNet \cite{wang2020model}       & 39.17          & 0.9885          & 30.24          & 0.9048          & 34.08          & 0.9532          & 33.04          & 0.9472          & 43.36          & 0.9831          \\
	\multicolumn{1}{c|}{}                                           & MPRNet \cite{zamir2021multi}      & 39.47          & 0.9825          & 30.67          & 0.9110          & 33.99          & 0.9590          & 33.10          & 0.9347          & 43.64              & 0.9844               \\
	\multicolumn{1}{c|}{}                                           & DualGCN \cite{fu2021rain}       & 40.73          & 0.9886          & 31.15          & 0.9125          & 34.37          & 0.9620          & 33.01          & 0.9489          & 44.18          & 0.9902          \\
	\multicolumn{1}{c|}{}                                           & SPDNet \cite{yi2021structure}       & 40.50          & 0.9875          & 31.28          & 0.9207          & 34.57          & 0.9560          & 33.15          & 0.9457          & 43.20          & 0.9871          \\ \hline
	\multicolumn{1}{c|}{\multirow{4}{*}{Transformer-based methods}} & Uformer \cite{wang2022uformer}    & 40.20          & 0.9860          & 30.80          & 0.9105          & 35.02          & 0.9621          &  33.95         & 0.9545          & 46.13          & 0.9913          \\
	\multicolumn{1}{c|}{}                                           & Restormer \cite{zamir2022restormer}    & \underline{40.99}          & \underline{0.9890}           & 32.00          & 0.9329          & \underline{35.29}          & \underline{0.9641}          & \underline{34.20}          & \underline{0.9571}          & \underline{47.98}          & 0.9921          \\
	\multicolumn{1}{c|}{}                                           & IDT \cite{xiao2022image}          & 40.74          & 0.9884          & \underline{32.10}          & \textbf{0.9344}          & 34.89          & 0.9623          & 33.84          & 0.9549          & 47.35          & \textbf{0.9930}          \\
	\multicolumn{1}{c|}{}                                           & \textbf{DRSformer} & \textbf{41.23} & \textbf{0.9894} & \textbf{32.18} & \underline{0.9330} & \textbf{35.38} & \textbf{0.9647} & \textbf{34.36} & \textbf{0.9590} & \textbf{48.53} & \underline{0.9924} \\ \hlinew{1.0pt}
\end{tabular}
}
\vspace{-5mm}
	\label{table1}	
\end{table*}

\subsection{Mixture of experts feature compensator}
To fill in the comprehensive faculty of integrating sparsity in the DRSformer, we further introduce MEFC to perform a unified co-exploration towards joint data and content sparsity.
Recalling the classical design of effective CNN models \cite{suganuma2019attention}, we elaborately select multiple sparse CNN operations to form parallel layers, dubbed as $\emph{experts}$, which involve an average pooling with receptive field of $3\times3$, separable convolution layers with kernel sizes of $1\times1$, $3\times3$, $5\times5$, $7\times7$, and dilated convolution layers with kernel sizes of $3\times3$, $5\times5$, $7\times7$.
Different from the conventional mixture of experts \cite{jacobs1991adaptive,ren2018gated}, our MEFC does not attach an external gating network.
Instead, we make the self-attention \cite{hu2018squeeze, kim2020restoring} become a switcher of different experts to adaptively select the importance of diverse representations depending on the inputs.
Given an input feature map $\mathbf{X}_{l-1} \in \mathbb{R}^{H \times W \times C}$, we first apply the channel-wise average to generate $C$-dimensional channel descriptor $\mathbf{z}_{c} \in \mathbbm{R}^{C} $:
\vspace{-1mm}
\begin{equation}
	\mathbf{z}_{c}=\frac{1}{H \times W} \sum_{i=1}^{H} \sum_{j=1}^{W} \mathbf{X}_{l-1}(i,j),
\vspace{-1mm}
\end{equation}
where $\mathbf{X}_{l-1}(i,j)$ is the $(y, x)$ position of the feature $\mathbf{X}_{l-1}$.
Then, the coefficient vector of each expert is allocated corresponding to the learnable weight matrices $\mathbf{W}_{1} \in \mathbbm{R}^{T \times C}$ and $\mathbf{W}_{2} \in \mathbbm{R}^{O \times T}$.
Here, $T$ is the dimension of the weight matrices. To avoid altering the sizes of its inputs and outputs, we zero pad the input feature maps computed by each expert. Finally, the output of the $l$-th MEFC is calculated by:
\vspace{-1mm}
\begin{equation}
	\begin{aligned}
		&\mathbf{T}_{l}= \mathbf{W}_{2} \sigma \left(\mathbf{W}_{1} \mathbf{z}_{c}\right), \\
		&\mathbf{X}_{l}=f_{1 \times 1}^c \left[\sum_{i=1}^O f_{exp}\left(\mathbf{X}_{l}, \mathbf{T}_{l}    \right)\right] + \mathbf{X}_{l-1},
	\end{aligned}
 \vspace{-1mm}
\end{equation}
where $f_{exp}$ and $O$ represent the expert operations and the number of experts, respectively. $f_{1 \times 1}^c$ represents $1 \times 1$ convolution, $\sigma(\cdot)$ is a ReLU function, and $[\cdot]$ is the channel-wise concatenation.
With this design, MEFC is now closely linked to the main STBs so that is able to adaptively remove the rainy effects of diverse appearances.

\begin{figure*}[t]
	\centering
	\includegraphics[width=1.0\textwidth]{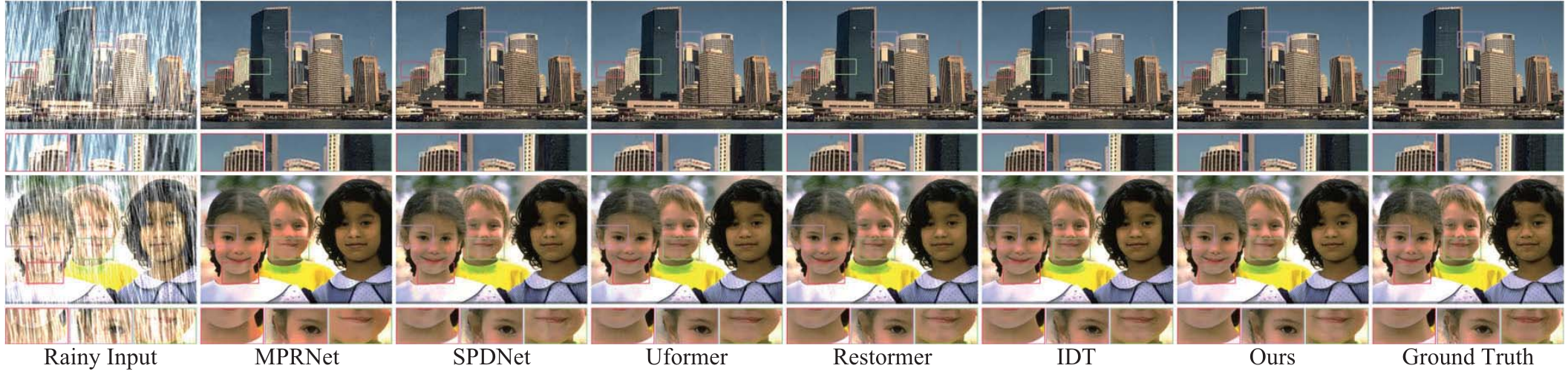}
	\vspace{-7mm}
	\caption{Visual quality comparison on the Rain200H dataset. Zooming in the figures offers a better view at the deraining capability.}
	\label{fig3}
	\vspace{-3mm}
\end{figure*}
\begin{figure*}[t]
	\centering
	\includegraphics[width=1.0\textwidth]{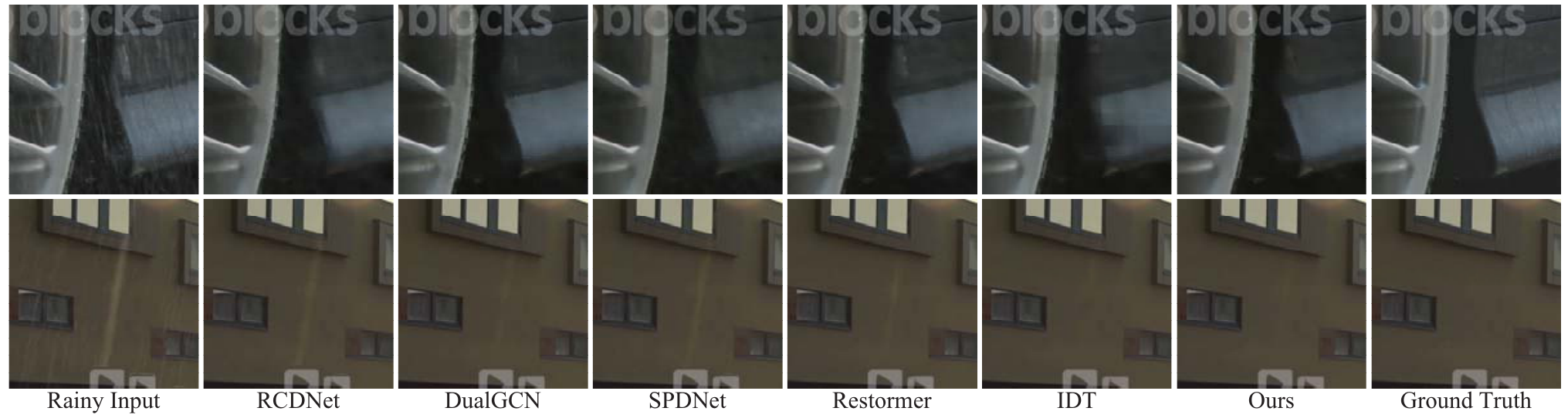}
	\vspace{-7mm}
	\caption{Visual quality comparison on the SPA-Data dataset. Zooming in the figures offers a better view at the deraining capability.}
	\label{fig4}
	\vspace{-5mm}
\end{figure*}
\begin{figure*}[t]
	\centering
	\includegraphics[width=1.0\textwidth]{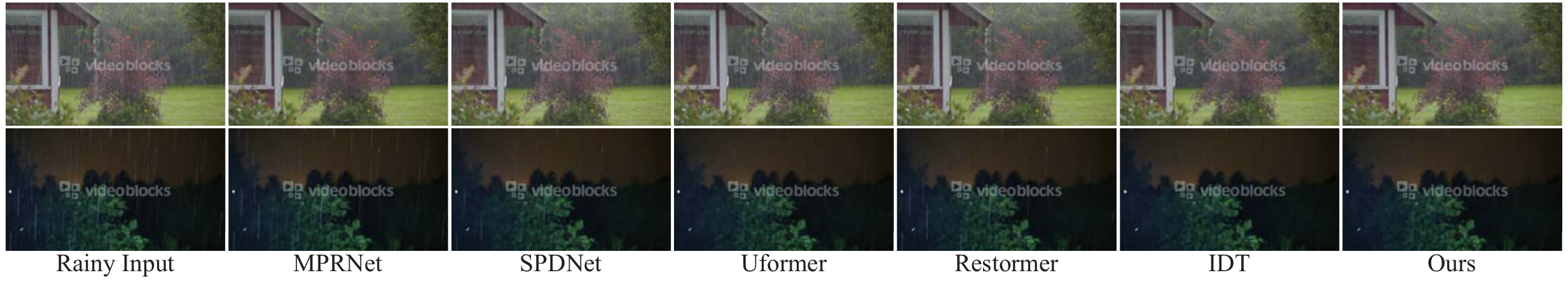}
	\vspace{-7mm}
	\caption{Visual quality comparison on real-world rainy images. Zooming in the figures offers a better view at the deraining capability.}
	\label{fig5}
	\vspace{-3mm}
\end{figure*}

\begin{table*}[t]\footnotesize
	\centering
	\caption{Comparison of quantitative results on real-world rainy images, and note that lower scores indicate better image quality.}
	\vspace{-3mm}
	\begin{tabular}{cccccccc}
		\hlinew{1.0pt}
		Methods        & Rainy Input & MPRNet \cite{zamir2021multi}        & SPDNet \cite{yi2021structure}        & Uformer \cite{wang2022uformer}       & Restormer \cite{zamir2022restormer}     & IDT \cite{xiao2022image}           & Ours    \\ \hline
		NIQE $\downarrow$ / BRISQUE $\downarrow$ & 5.829 / 33.129 & 4.740 / 32.018 & 4.422 / 26.173 & 4.833 / 28.106 & 5.005 / 34.036 & 4.238 / 25.573 & \textbf{4.095} / \textbf{22.730} \\ \hlinew{1.0pt}
	\end{tabular}
	\vspace{-3mm}
	\label{table2}	
\end{table*}

\vspace{-1mm}
\section{Experiments and Analysis}
\vspace{-1mm}
\subsection{Experimental settings}
{\flushleft \textbf{Datasets}.}
We implement deraining experiments on multiple public benchmarks, including Rain200L/H \cite{yang2017deep}, DID-Data \cite{zhang2018density} and DDN-Data \cite{fu2017removing}.
Rain200L and Rain200H contain 1,800 synthetic rainy images for training and 200 ones for testing.
DID-Data and DDN-Data consist of 12,000 and 12,600 synthetic images with different rain directions and density levels. There are 1,200 and 1,400 rainy images for testing.
In addition, we also evaluate our method using a large-scale real-world dataset, \emph{i.e}, SPA-Data \cite{wang2019spatial}, containing 638,492 image pairs for training and 1,000 testing ones.

\vspace{-2mm}
{\flushleft \textbf{Comparison methods}.}
We compare our DRSformer with two prior-based models (DSC \cite{luo2015removing} and GMM \cite{li2016rain}), CNN-based methods (DDN \cite{fu2017removing}, RESCAN \cite{li2018recurrent}, PReNet \cite{ren2019progressive}, MSPFN \cite{jiang2020multi}, RCDNet~\cite{wang2020model}, MPRNet \cite{zamir2021multi}, DualGCN \cite{fu2021rain}, and SPDNet \cite{yi2021structure}), and recent Transformer-based methods (Uformer \cite{wang2022uformer}, Restormer \cite{zamir2022restormer}, and IDT \cite{xiao2022image}).
For recent representative methods (DualGCN, SPDNet, Restormer and IDT), we retrain their models provided by the authors if no pretrained models are provided, otherwise we evaluate them with their online codes for fair comparisons.
For other approaches, we refer to some reported results in \cite{fu2023continual, xiao2022image}.
\vspace{-2mm}
{\flushleft \textbf{Evaluation metrics}.}
We adopt PSNR \cite{PSNR} and SSIM \cite{wang2004image} as the evaluation metrics for the above benchmarks.
Following previous deraining methods \cite{jiang2020multi,fu2023continual}, we calculate PSNR and SSIM metrics in Y channel of YCbCr space.
For the rainy images without ground truth images, we use the non-reference metrics including NIQE \cite{mittal2012making} and BRISQUE \cite{mittal2012no}.

\vspace{-2mm}
{\flushleft \textbf{Training details}.}
In our model, $\left\{N_0, N_1, N_2, N_3, N_4\right\}$ are set to $\{4,4,6,6,8\}$, and the number of attention heads for four STBs of the same level is set to $\{1,2,4,8\}$. The initial channel $C$ is 48 and the expanding ratio is set to 2. In terms of MEFC, we set $O = 8$ for the number of experts and $T = 32$ for the weight matrix. Note that we do not use MEFC for training Rain200L and SPA-Data, because their rain streaks are less complex and easier to learn. In terms of STB, the sparseness $[\Delta_{1}, \Delta_{2}]$ in TKSA is set to $[\frac{1}{2}, \frac{4}{5}]$, and the channel expansion factor $r$ in MSFN is set to 2.66. During training, we use AdamW optimizer with batch size of 8 and patch size of 128 for total 300K iterations. The initial learning rate is fixed as $1 \times 10^{-4}$ for 92K iterations, and then decreased to $1 \times 10^{-6}$ for 208K iterations with the cosine annealing scheme \cite{loshchilov2016sgdr}. For data augmentation, vertical and horizontal flips are randomly applied. The entire framework is performed on the PyTorch with 4 NVIDIA GeForce RTX 3090 GPUs, which works in an end-to-end learning fashion without costly large-scale pretraining \cite{chen2021pre}.

\vspace{-1mm}
\subsection{Comparisons with the state-of-the-arts}
\vspace{-1mm}
{\flushleft\textbf{Synthetic datasets}.}
The quantitative evaluations on different benchmark datasets are reported in Table~ \ref{table1}.
As shown, we can note that our proposed method outperforms all the other derainers, especially on PSNR, \emph{e.g.}, DRSformer surpasses the concurrent approach IDT by 0.4 dB on average.
Compared with previous CNN-based models, this progress can be much more obvious.
The notable increasing scores on the DID-Data and DDN-Data benchmarks reveal that our method can properly handle diverse types of spatially-varying rain streaks.
For convincing evidence, we show the visual quality comparison between samples generated by recent approaches in Figure \ref{fig3}.
The pure CNN-based models, \emph{e.g.}, MPRNet and SPDNet, fail to restore clear images in heavy rainy scenarios.
It can be seen that the results of all computing Transformer-based methods are flawed in terms of detail and texture recovery.
Unfortunately, IDT even introduces considerable boundary artifacts.
Thanks to the developed sparse attention with top-\emph{k} selection, our method can generate high-quality deraining results, which are more consistent with that of the ground truth.
%
\vspace{-2mm}
{\flushleft\textbf{Real-world datasets}.}
We further conduct experiments on the SPA-Data benchmark dataset, and corresponding results are provided in the last column of Table~\ref{table1}.
As expected, our model continues to achieve the highest PSNR/SSIM value, exhibiting the superior of DRSformer in terms of deraining performance and generalization.
The visual quality comparison can be observed in Figure \ref{fig4}.
In contrast, our method significantly competes with others in removing most rain streaks while preserving truthful image structures.
In order to further validate the effectiveness of DRSformer, we also randomly choose 20 real rainy images without ground truths from Internet-Data \cite{wang2019spatial} to perform another evaluation.
As displayed in Table~\ref{table2}, our net gets the lower NIQE and BRISQUE values, which means a high-quality output with clearer content and better perceptual quality against other comparative models on the real rainy scenarios.
Through qualitative quality comparison in Figure \ref{fig5}, most deep models are sensitive to spatially-long rain streaks and leave some apparent rain effects.
On the contrary, our net successfully removes most rain perturbation and owns a visual pleasant recovery effect, which implies that it can well generalize to unseen real-world data types.

\begin{figure}[!t]
	\centering
	\includegraphics[width=1.0\columnwidth]{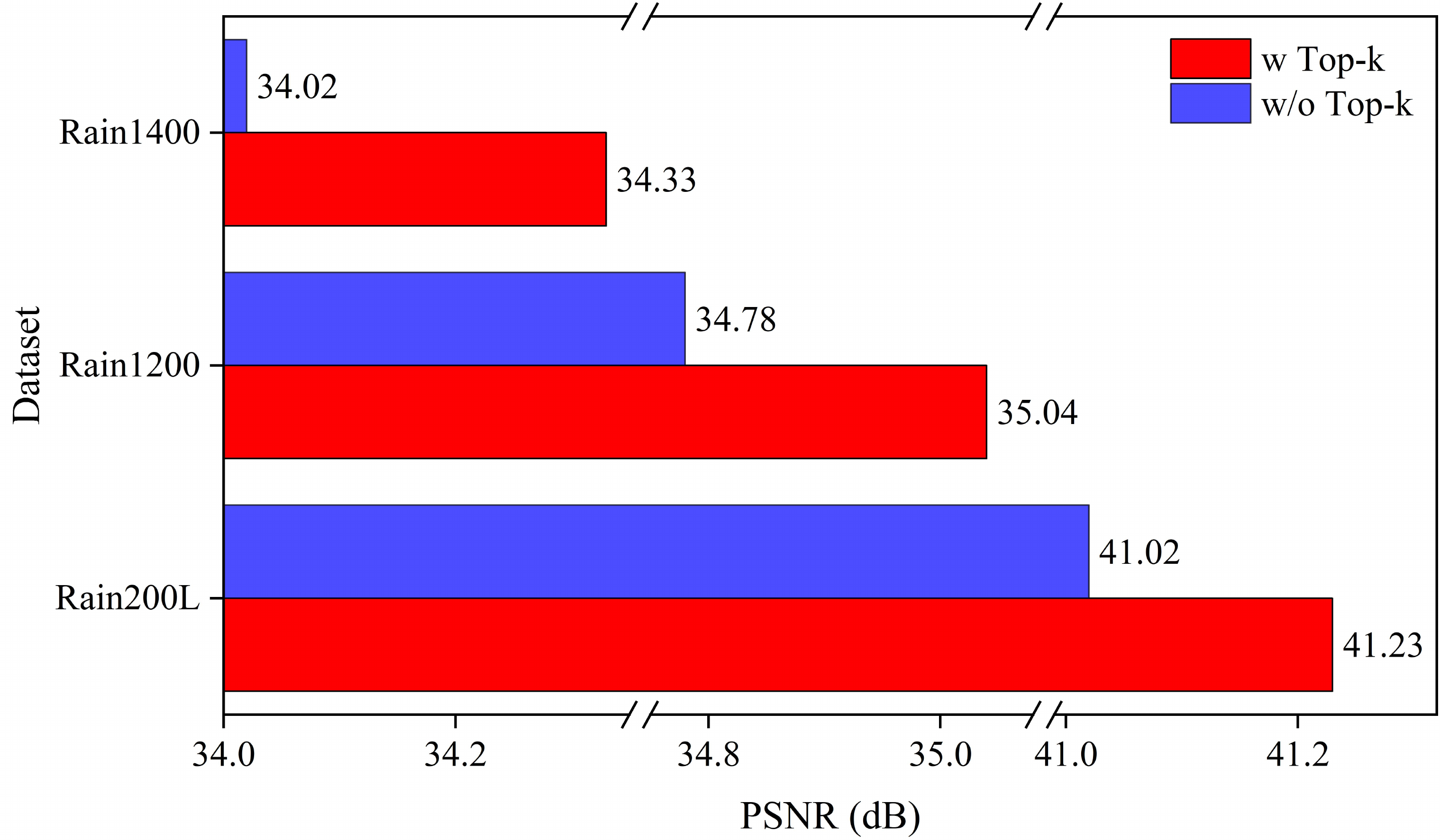}
	\vspace{-6mm}
	\caption{Ablation analysis for top-\emph{k} selection on the benchmarks.}
	\label{fig6}
	\vspace{-4mm}
\end{figure}

\vspace{-1mm}
\subsection{Ablation studies}
\vspace{-1mm}
\label{4.3}
{\flushleft\textbf{Effectiveness of Top-\emph{k} selection}.}
To examine the effect of the top-\emph{k} selection in the TKSA, we present the deraining results of TKSA w/o top-\emph{k} in Figure \ref{fig6}.
We can see that the PSNR values of the derained images by the method without using the top-\emph{k} selection are lower than those by the method using the top-\emph{k} selection.
In addition, we also note that each element of the self-attention matrix $\operatorname{softmax}\left(\frac{\mathbf{Q} \mathbf{K}^{\top}}{\lambda}\right)$~ in Eq.~\eqref{4} is non-negative and the summation of the elements from each row of $\operatorname{softmax}\left(\frac{\mathbf{Q} \mathbf{K}^{\top}}{\lambda}\right)$ is 1.
Thus, applying the self-attention matrix $\operatorname{softmax}\left(\frac{\mathbf{Q} \mathbf{K}^{\top}}{\lambda}\right)$ to $\mathbf{V}$ will remove the high-frequency information of $\mathbf{V}$, which lead to over-smoothed results.

To understand the effect of such top-\emph{k} selection, we further use high-pass filtering (HPF) to visualize learned features in Figure \ref{fig8}.
Compared to standard self-attention operation (w/o top-\emph{k}), our strategy can better help reconstruct finer-detail feature and improve potential restoration quality.
As the nearby pixels tend to be more similar than others, top-\emph{k} selection operator helps to reduce the irrelevant context from long-range pixel dependency.
This step of selection allows the smaller similarity weights (from a part of long-range feature interactions) to be discarded in the procedure of self-attention calculation, thus facilitating more accurate representation for achieving high-quality output.

\begin{figure}[!t]
	\centering
	\includegraphics[width=1.0\columnwidth]{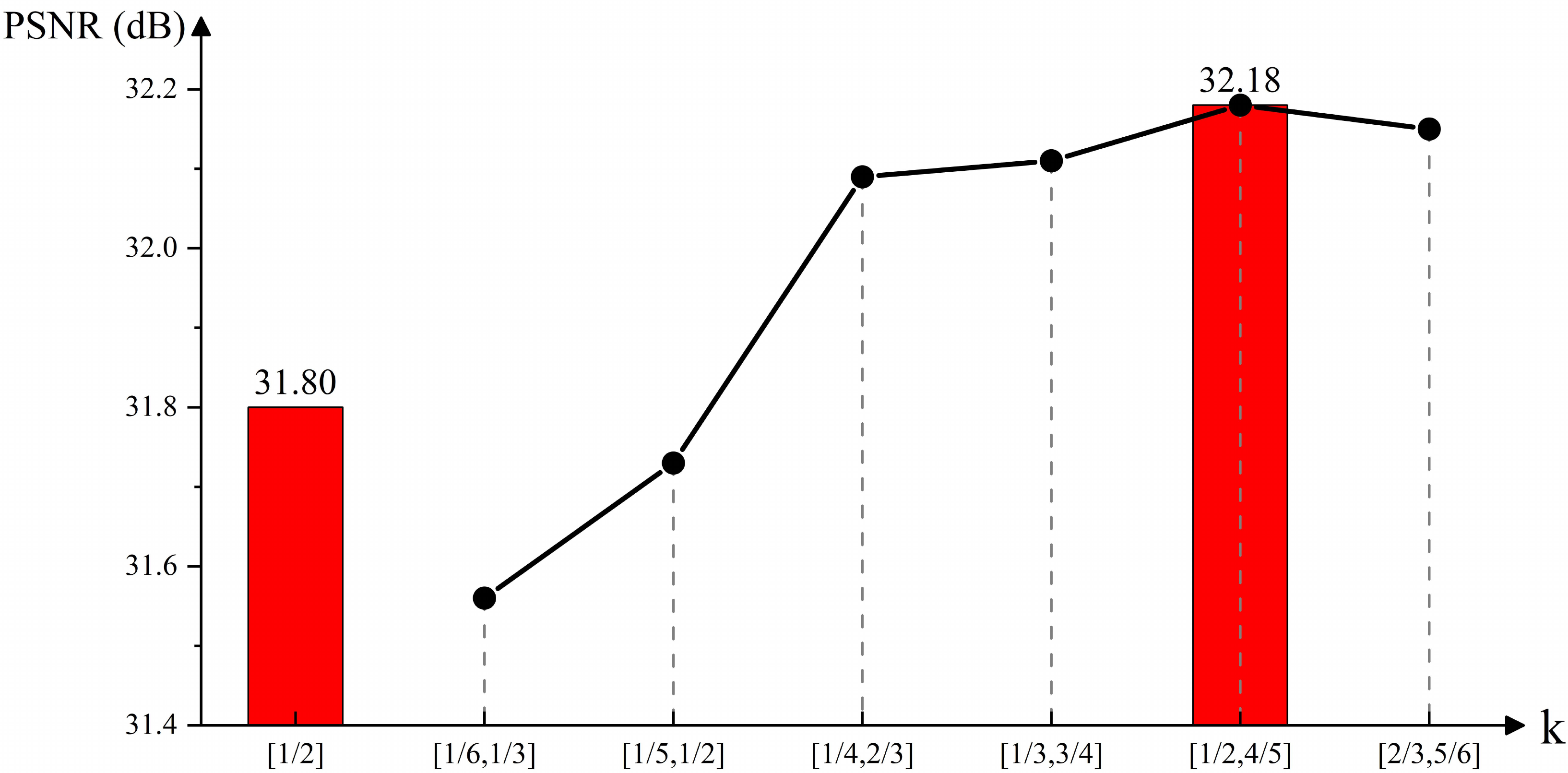}
	\vspace{-6mm}
	\caption{Ablation analysis for different number \emph{k} in the TKSA.}
	\label{fig7}
	\vspace{-1mm}
\end{figure}

\begin{figure}[!t]
	\centering
	\includegraphics[width=1.0\columnwidth]{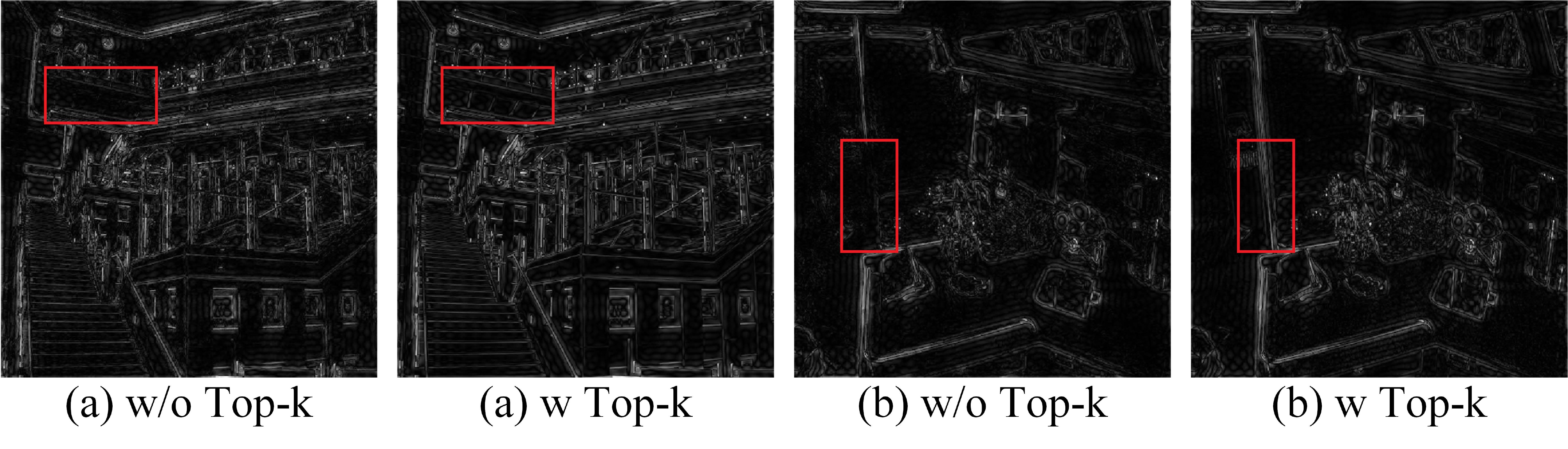}
	\vspace{-6mm}
	\caption{Visualization of feature maps. Our proposed top-\emph{k} selection can effectively leverage pixel-dependent properties of image structure and generate more precise high-frequency details.}
	\label{fig8}
	\vspace{-4mm}
\end{figure}

\vspace{-2mm}
{\flushleft\textbf{Effect of the number of \emph{k}}.}
The key parameter for our proposed TKSA is \emph{k}, and its influence is investigated in Figure \ref{fig7}.
We note that the optimal choice of \emph{k} determines the boundary control of the sparsity rate. If $k$ is manually set to a single value, such as $\frac{1}{2}$, we notice that its performance is sensitive to $k$.
To avoid an exhaustive search, we set a controllable interval range for $k$ to dynamically learn the most contribute score.
When \emph{k} is too small, we find that the performance will undoubtedly decline sharply due to insufficient global information aggregation.
The best result can achieve 32.18 dB when $[\Delta_{1}, \Delta_{2}]$ in the TKSA is assigned to $[\frac{1}{2}, \frac{4}{5}]$.
As \emph{k} continues to increase, the final deraining performance gradually decreases due to the introduction of irrelevant and useless features.

\begin{figure*}[t]
	\centering
	\includegraphics[width=1.0\textwidth]{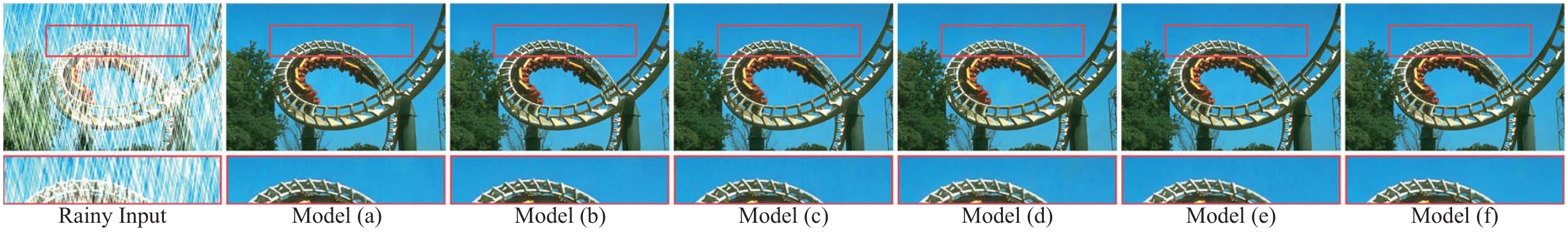}
	\vspace{-7mm}
	\caption{Ablation qualitative comparison for different variants of DRSformer. The models (a-f) are consistent with the settings in Table \ref{table4}.}
	\label{fig9}
	\vspace{-5mm}
\end{figure*}

\vspace{-3mm}
{\flushleft\textbf{Effectiveness of MSFN}.}
To evaluate the effectiveness of the proposed MSFN, we compare it with three baselines: (1) conventional feed-forward network (FN) \cite{dosovitskiy2020image}, (2) Dconv feed-forward network (DFN) \cite{li2021localvit}, and (3) gated-Dconv feed-forward network (GDFN) \cite{zamir2022restormer}.
The quantitative analysis results on Rain200H are listed in Table~\ref{table3}.
Although GDFN introduces a gating mechanism in two same-scale depth-wise convolution streams to bring performance advantages, it still neglects the multi-scale knowledge for deraining.
By adding local feature extraction and fusion at different scales, the MSFN can indeed better boost the performance, and achieve PSNR gain of 0.21 dB over GDFN.

\vspace{-1mm}
{\flushleft\textbf{Effectiveness of MEFC}.}
To evaluate the effectiveness of MEFC, we perform experiments based on different model variants in Table~\ref{table4}.
Compared to the baseline model (a), MEFC provides additional performance benefits thanks to auxiliary data sparsity.
In addition, we observe that MEFCs at different locations of the network pipeline have specific impacts on the restoration performance.
Indeed, we also analyze the effect of the different numbers of experts in each MEFC.
When using single expert model (d), the performance is dramatically degraded compared with our multi-expert model (f).
Unlike setting all experts to the same structure \cite{kim2020restoring}, our multi-expert formulation is more diverse, which brings its gains to the performance due to different receptive fields and disparate CNN operations.
Through the zoomed boxes in Figure \ref{fig9}, the recovered results of the model with all the above components tend to be clearer since it enables more diverse features to be fully used during the restoration process.
All in all, our model (f) performs better than the other possible configurations, which indicates that each design strategy that we consider has its own contribution to the final performance of DRSformer.

\begin{table}[t]
	\centering
	\caption{Ablation study for different feed-forward networks.}
	\vspace{-3mm}
	\resizebox{1.0\columnwidth}{!}{
	\begin{tabular}{ccccc}
	\hlinew{1.0pt}
	Models       & FN \cite{dosovitskiy2020image}           & DFN \cite{li2021localvit}          & GDFN \cite{zamir2022restormer}         & MSFN                   \\ \hline
	PSNR / SSIM & 31.84 / 0.9275 & 31.88 / 0.9279 & 31.97 / 0.9286 & \textbf{32.18} / \textbf{0.9330} \\ \hlinew{1.0pt}
\end{tabular}
	}	
	\label{table3}
	\vspace{-4mm}
\end{table}

\begin{table}[t]\small
	\centering
	\caption{Ablation study for different variants of our DRSformer. MEFC-1 and MEFC-2 denote MEFC in early and final stages.}
	\vspace{-3mm}
	\resizebox{1.0\columnwidth}{!}{
	\begin{tabular}{cccccc}
	\hlinew{1.0pt}
	Models & MEFC-1 & STBs & MEFC-2 & Experts & PSNR / SSIM   \\ \hline
	(a)    &        & $\checkmark$    &        & 0       & 32.03 / 0.9308 \\
	(b)    & $\checkmark$       & $\checkmark$    &       & 8       & 32.01 / 0.9311 \\
	(c)    &       & $\checkmark$    & $\checkmark$       & 8       & 32.07 / 0.9328 \\
	(d)    & $\checkmark$      & $\checkmark$    & $\checkmark$      & 1       & 32.06 / 0.9316 \\
	(e)    & $\checkmark$      & $\checkmark$    & $\checkmark$      & 4       & 32.14 / 0.9325 \\
	(f)    & $\checkmark$      & $\checkmark$    & $\checkmark$      & 8       & \textbf{32.18} / \textbf{0.9330} \\ \hlinew{1.0pt}
    \end{tabular}
	}	
	\label{table4}
	\vspace{-5mm}
\end{table}

\subsection{Closely-related methods}
\begin{figure}[!t]
	\centering 	
	\begin{subfigure}[t]{0.24\columnwidth}
		\centering
		\includegraphics[width=\columnwidth]{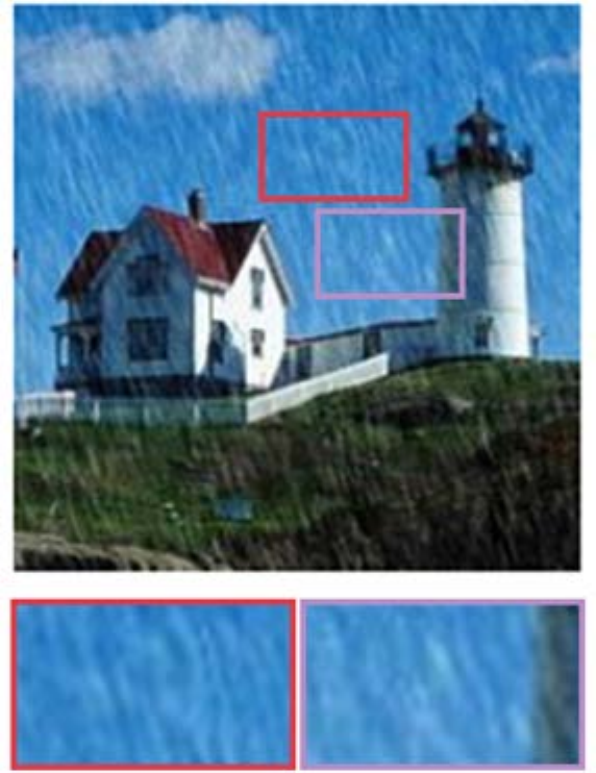}
		\caption{Rainy Input}
	\end{subfigure}
	\begin{subfigure}[t]{0.24\columnwidth}
		\centering
		\includegraphics[width=\columnwidth]{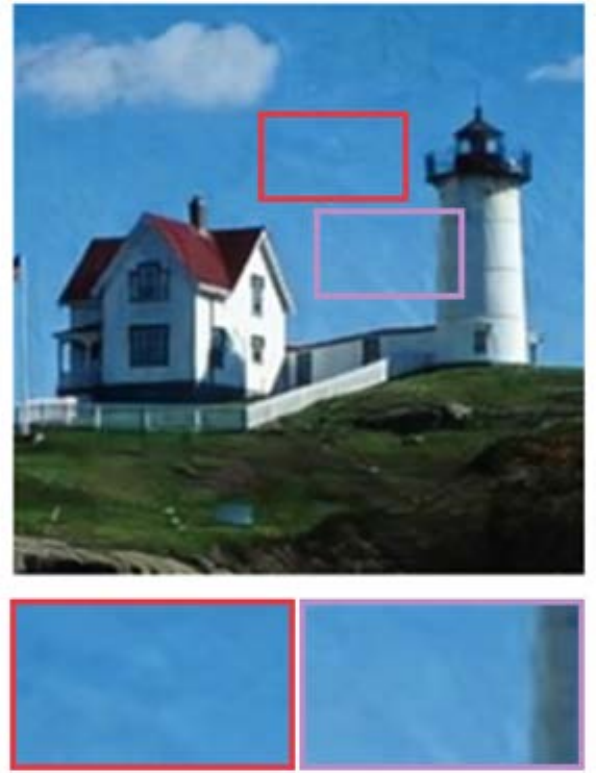}
		\caption{KiT \cite{lee2022knn}}
	\end{subfigure}
	\begin{subfigure}[t]{0.24\columnwidth}
	\centering
	\includegraphics[width=\columnwidth]{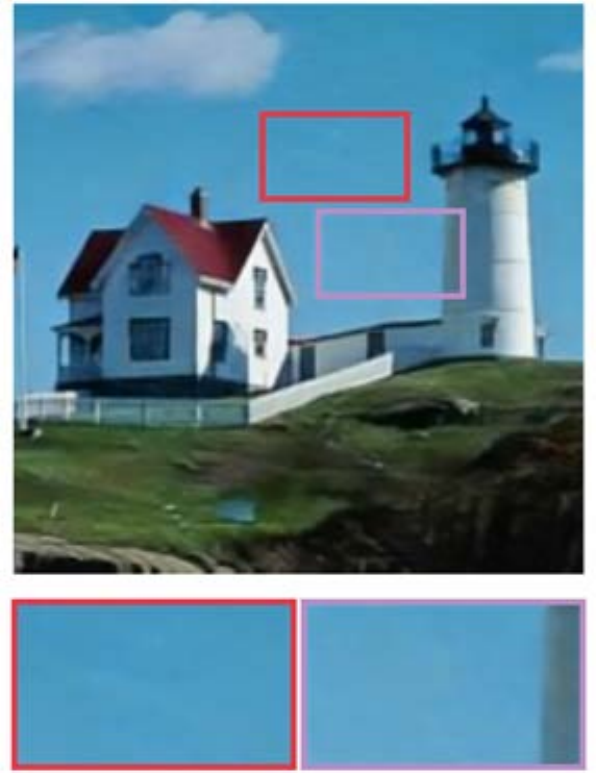}
	\caption{KVT \cite{wang2022kvt}}
    \end{subfigure}	
    \begin{subfigure}[t]{0.24\columnwidth}
    \centering
    \includegraphics[width=\columnwidth]{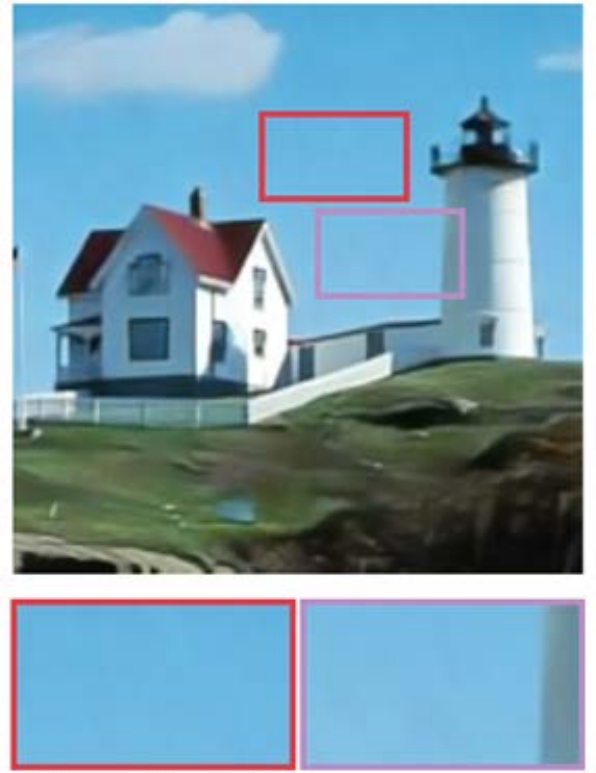}
    \caption{Ours}
    \end{subfigure}
	\vspace{-3mm}
	\caption{Comparison results with closely-related methods.}
	\label{fig10}
	\vspace{-6mm}
\end{figure}

We note that the recent method \cite{lee2022knn} proposes a \emph{k}-NN image Transformer (KiT) to solve image restoration by aggregating \emph{k} similar patches with the pair-wise local attention.
Compared with KiT that employs complex locality sensitive hashing that cannot ensure sufficient global interaction, our simple but effective top-\emph{k} selection mechanism not only enjoys the locality but also empowers the ability of global relation mining.
As the code of KiT is not available, we refer to the results of their paper.
Figure \ref{fig10} shows qualitative comparisons trained on the Rain800 \cite{zhang2019image}.
We can see that KiT tends to blur the contents and cause color distortion.
In contrast, our method leads to better deraining results.

In addition, we also note that \cite{wang2022kvt} recently designs the \emph{k}-NN attention to enhance the representation ability of vision Transformers by selecting the top-\emph{k} similar tokens.
Different from KVT \cite{wang2022kvt}, which implements top-\emph{k} selection in the spatial dimension, our operator is more efficient in computing sparse attention across channels.
Furthermore, the sparsity level of $k$ in our proposed TKSA is dynamically learnable, rather than the fixed setting in \cite{wang2022kvt}.
Here, we adopt \emph{k}-NN attention in KVT to replace our TKSA for comparison.
To ensure the fair comparison, the same training settings are kept for model testing.
As shown in Figure \ref{fig10} (c) and (d), our method can generate a clearer image.

\vspace{-1mm}
\section{Concluding Remarks}
\vspace{-1mm}
We have presented an effective sparse Transformer network, called DRSformer, to solve image deraining.
Based on the observation that vanilla self-attention in Transformer may suffer from the global interaction of irrelevant information, we develop the top-\emph{k} sparse attention to keep the most useful self-attention values for better feature aggregation.
To facilitate the aggregated features for removing rain, we develop a mixed-scale feed-forward network to better explore multi-scale representations.
Furthermore, the mixture of experts feature compensator is introduced to the model to provide collaborative refinement for the sparse Transformer backbone, so that the fine details of the reconstructed image is preserved.
Experimental results show that our DRSformer performs favorably against state-of-the-art methods.
\vspace{-4.5mm}
{\flushleft\textbf{Limitations}.}
Our proposed method aims to further boost image deraining performance, but there are limitations in the model efficiency. Specifically, our model requires 33.7 Million parameters and costs 242.9G FLOPs on one image with size of $256 \times 256$. We will apply the pruning or distillation scheme in our model to maintain the original deraining performance while achieving credible model compression.

\vspace{-2mm}
{\flushleft\textbf{Acknowledgements}.} This work has been partly supported by the National Key R\&D Program of China (No. 2018AAA0102001), the National Natural Science Foundation of China (Nos. U22B2049, U19B2040, 61922043, 61872421, 62272230), and the Fundamental Research Funds for the Central Universities (No. 30920041109).

{\small
\bibliographystyle{ieee_fullname}
\bibliography{egbib}

\begin{thebibliography}{10}\itemsep=-1pt

\bibitem{carion2020end}
Nicolas Carion, Francisco Massa, Gabriel Synnaeve, Nicolas Usunier, Alexander
  Kirillov, and Sergey Zagoruyko.
\newblock End-to-end object detection with transformers.
\newblock In {\em ECCV}, pages 213--229, 2020.

\bibitem{chen2021pre}
Hanting Chen, Yunhe Wang, Tianyu Guo, Chang Xu, Yiping Deng, Zhenhua Liu, Siwei
  Ma, Chunjing Xu, Chao Xu, and Wen Gao.
\newblock Pre-trained image processing transformer.
\newblock In {\em CVPR}, pages 12299--12310, 2021.

\bibitem{chen2021chasing}
Tianlong Chen, Yu Cheng, Zhe Gan, Lu Yuan, Lei Zhang, and Zhangyang Wang.
\newblock Chasing sparsity in vision transformers: An end-to-end exploration.
\newblock {\em Advances in Neural Information Processing Systems},
  34:19974--19988, 2021.

\bibitem{chen2022unpaired}
Xiang Chen, Jinshan Pan, Kui Jiang, Yufeng Li, Yufeng Huang, Caihua Kong,
  Longgang Dai, and Zhentao Fan.
\newblock Unpaired deep image deraining using dual contrastive learning.
\newblock In {\em CVPR}, pages 2017--2026, 2022.

\bibitem{chen2023hybrid}
Xiang Chen, Jinshan Pan, Jiyang Lu, Zhentao Fan, and Hao Li.
\newblock Hybrid cnn-transformer feature fusion for single image deraining.
\newblock In {\em AAAI}, 2023.

\bibitem{correia2019adaptively}
Gon{\c{c}}alo~M Correia, Vlad Niculae, and Andr{\'e}~FT Martins.
\newblock Adaptively sparse transformers.
\newblock {\em arXiv preprint arXiv:1909.00015}, 2019.

\bibitem{dosovitskiy2020image}
Alexey Dosovitskiy, Lucas Beyer, Alexander Kolesnikov, Dirk Weissenborn,
  Xiaohua Zhai, Thomas Unterthiner, Mostafa Dehghani, Matthias Minderer, Georg
  Heigold, Sylvain Gelly, et~al.
\newblock An image is worth 16x16 words: Transformers for image recognition at
  scale.
\newblock In {\em ICLR}, 2021.

\bibitem{fu2017removing}
Xueyang Fu, Jiabin Huang, Delu Zeng, Yue Huang, Xinghao Ding, and John Paisley.
\newblock Removing rain from single images via a deep detail network.
\newblock In {\em CVPR}, pages 3855--3863, 2017.

\bibitem{fu2021rain}
Xueyang Fu, Qi Qi, Zheng-Jun Zha, Yurui Zhu, and Xinghao Ding.
\newblock Rain streak removal via dual graph convolutional network.
\newblock In {\em AAAI}, pages 1352--1360, 2021.

\bibitem{fu2023continual}
Xueyang Fu, Jie Xiao, Yurui Zhu, Aiping Liu, Feng Wu, and Zheng-Jun Zha.
\newblock Continual image deraining with hypergraph convolutional networks.
\newblock {\em IEEE TPAMI}, 2023.

\bibitem{fu2022sparsett}
Zhihong Fu, Zehua Fu, Qingjie Liu, Wenrui Cai, and Yunhong Wang.
\newblock Sparsett: Visual tracking with sparse transformers.
\newblock {\em arXiv preprint arXiv:2205.03776}, 2022.

\bibitem{gu2017joint}
Shuhang Gu, Deyu Meng, Wangmeng Zuo, and Lei Zhang.
\newblock Joint convolutional analysis and synthesis sparse representation for
  single image layer separation.
\newblock In {\em ICCV}, 2017.

\bibitem{guo2022image}
Chun-Le Guo, Qixin Yan, Saeed Anwar, Runmin Cong, Wenqi Ren, and Chongyi Li.
\newblock Image dehazing transformer with transmission-aware 3d position
  embedding.
\newblock In {\em CVPR}, pages 5812--5820, 2022.

\bibitem{hu2018squeeze}
Jie Hu, Li Shen, and Gang Sun.
\newblock Squeeze-and-excitation networks.
\newblock In {\em CVPR}, pages 7132--7141, 2018.

\bibitem{hu2019depth}
Xiaowei Hu, Chi-Wing Fu, Lei Zhu, and Pheng-Ann Heng.
\newblock Depth-attentional features for single-image rain removal.
\newblock In {\em CVPR}, pages 8022--8031, 2019.

\bibitem{huang2021memory}
Huaibo Huang, Aijing Yu, and Ran He.
\newblock Memory oriented transfer learning for semi-supervised image
  deraining.
\newblock In {\em CVPR}, pages 7732--7741, 2021.

\bibitem{jacobs1991adaptive}
Robert~A Jacobs, Michael~I Jordan, Steven~J Nowlan, and Geoffrey~E Hinton.
\newblock Adaptive mixtures of local experts.
\newblock {\em Neural computation}, 3(1):79--87, 1991.

\bibitem{jiang2022magic}
Kui Jiang, Zhongyuan Wang, Chen Chen, Zheng Wang, Laizhong Cui, and Chia-Wen
  Lin.
\newblock Magic elf: Image deraining meets association learning and
  transformer.
\newblock In {\em ACM MM}, 2022.

\bibitem{jiang2020multi}
Kui Jiang, Zhongyuan Wang, Peng Yi, Chen Chen, Baojin Huang, Yimin Luo, Jiayi
  Ma, and Junjun Jiang.
\newblock Multi-scale progressive fusion network for single image deraining.
\newblock In {\em CVPR}, pages 8346--8355, 2020.

\bibitem{kang2011automatic}
Li-Wei Kang, Chia-Wen Lin, and Yu-Hsiang Fu.
\newblock Automatic single-image-based rain streaks removal via image
  decomposition.
\newblock {\em IEEE TIP}, 21(4):1742--1755, 2011.

\bibitem{kim2020restoring}
Sijin Kim, Namhyuk Ahn, and Kyung-Ah Sohn.
\newblock Restoring spatially-heterogeneous distortions using mixture of
  experts network.
\newblock In {\em ACCV}, 2020.

\bibitem{lee2022knn}
Hunsang Lee, Hyesong Choi, Kwanghoon Sohn, and Dongbo Min.
\newblock Knn local attention for image restoration.
\newblock In {\em CVPR}, pages 2139--2149, 2022.

\bibitem{li2018recurrent}
Xia Li, Jianlong Wu, Zhouchen Lin, Hong Liu, and Hongbin Zha.
\newblock Recurrent squeeze-and-excitation context aggregation net for single
  image deraining.
\newblock In {\em ECCV}, pages 254--269, 2018.

\bibitem{li2016rain}
Yu Li, Robby~T Tan, Xiaojie Guo, Jiangbo Lu, and Michael~S Brown.
\newblock Rain streak removal using layer priors.
\newblock In {\em CVPR}, pages 2736--2744, 2016.

\bibitem{li2021localvit}
Yawei Li, Kai Zhang, Jiezhang Cao, Radu Timofte, and Luc Van~Gool.
\newblock Localvit: Bringing locality to vision transformers.
\newblock {\em arXiv preprint arXiv:2104.05707}, 2021.

\bibitem{liang2022drt}
Yuanchu Liang, Saeed Anwar, and Yang Liu.
\newblock Drt: A lightweight single image deraining recursive transformer.
\newblock In {\em CVPRW}, pages 589--598, 2022.

\bibitem{liu2021unpaired}
Yang Liu, Ziyu Yue, Jinshan Pan, and Zhixun Su.
\newblock Unpaired learning for deep image deraining with rain direction
  regularizer.
\newblock In {\em ICCV}, pages 4753--4761, 2021.

\bibitem{liu2021swin}
Ze Liu, Yutong Lin, Yue Cao, Han Hu, Yixuan Wei, Zheng Zhang, Stephen Lin, and
  Baining Guo.
\newblock Swin transformer: Hierarchical vision transformer using shifted
  windows.
\newblock In {\em ICCV}, pages 10012--10022, 2021.

\bibitem{loshchilov2016sgdr}
Ilya Loshchilov and Frank Hutter.
\newblock Sgdr: Stochastic gradient descent with warm restarts.
\newblock In {\em ICLR}, 2016.

\bibitem{luo2015removing}
Yu Luo, Yong Xu, and Hui Ji.
\newblock Removing rain from a single image via discriminative sparse coding.
\newblock In {\em ICCV}, pages 3397--3405, 2015.

\bibitem{mei2021image}
Yiqun Mei, Yuchen Fan, and Yuqian Zhou.
\newblock Image super-resolution with non-local sparse attention.
\newblock In {\em CVPR}, pages 3517--3526, 2021.

\bibitem{mittal2012no}
Anish Mittal, Anush~Krishna Moorthy, and Alan~Conrad Bovik.
\newblock No-reference image quality assessment in the spatial domain.
\newblock {\em IEEE TIP}, 21(12):4695--4708, 2012.

\bibitem{mittal2012making}
Anish Mittal, Rajiv Soundararajan, and Alan~C Bovik.
\newblock Making a “completely blind” image quality analyzer.
\newblock {\em IEEE SPL}, 20(3):209--212, 2012.

\bibitem{PSNR}
Huynh-Thu Q. and Ghanbari M.
\newblock Scope of validity of psnr in image/video quality assessment.
\newblock {\em Electronics Letters}, 44(13):800--801, 2008.

\bibitem{qin2021etdnet}
Qin Qin, Jingke Yan, Qin Wang, Xin Wang, Minyao Li, and Yuqing Wang.
\newblock Etdnet: An efficient transformer deraining model.
\newblock {\em IEEE Access}, 9:119881--119893, 2021.

\bibitem{ren2019progressive}
Dongwei Ren, Wangmeng Zuo, Qinghua Hu, Pengfei Zhu, and Deyu Meng.
\newblock Progressive image deraining networks: A better and simpler baseline.
\newblock In {\em CVPR}, pages 3937--3946, 2019.

\bibitem{ren2018gated}
Wenqi Ren, Lin Ma, Jiawei Zhang, Jinshan Pan, Xiaochun Cao, Wei Liu, and
  Ming-Hsuan Yang.
\newblock Gated fusion network for single image dehazing.
\newblock In {\em CVPR}, pages 3253--3261, 2018.

\bibitem{roy2021efficient}
Aurko Roy, Mohammad Saffar, Ashish Vaswani, and David Grangier.
\newblock Efficient content-based sparse attention with routing transformers.
\newblock {\em Transactions of the Association for Computational Linguistics},
  9:53--68, 2021.

\bibitem{suganuma2019attention}
Masanori Suganuma, Xing Liu, and Takayuki Okatani.
\newblock Attention-based adaptive selection of operations for image
  restoration in the presence of unknown combined distortions.
\newblock In {\em CVPR}, pages 9039--9048, 2019.

\bibitem{vaswani2017attention}
Ashish Vaswani, Noam Shazeer, Niki Parmar, Jakob Uszkoreit, Llion Jones,
  Aidan~N Gomez, {\L}ukasz Kaiser, and Illia Polosukhin.
\newblock Attention is all you need.
\newblock {\em Advances in neural information processing systems}, 30, 2017.

\bibitem{wang2020dcsfn}
Cong Wang, Xiaoying Xing, Yutong Wu, Zhixun Su, and Junyang Chen.
\newblock Dcsfn: Deep cross-scale fusion network for single image rain removal.
\newblock In {\em ACM MM}, pages 1643--1651, 2020.

\bibitem{wang2022nformer}
Haochen Wang, Jiayi Shen, Yongtuo Liu, Yan Gao, and Efstratios Gavves.
\newblock Nformer: Robust person re-identification with neighbor transformer.
\newblock In {\em CVPR}, pages 7297--7307, 2022.

\bibitem{wang2020model}
Hong Wang, Qi Xie, Qian Zhao, and Deyu Meng.
\newblock A model-driven deep neural network for single image rain removal.
\newblock In {\em CVPR}, pages 3103--3112, 2020.

\bibitem{wang2022kvt}
Pichao Wang, Xue Wang, Fan Wang, Ming Lin, Shuning Chang, Wen Xie, Hao Li, and
  Rong Jin.
\newblock Kvt: k-nn attention for boosting vision transformers.
\newblock In {\em ECCV}, 2022.

\bibitem{wang2019spatial}
Tianyu Wang, Xin Yang, Ke Xu, Shaozhe Chen, Qiang Zhang, and Rynson~WH Lau.
\newblock Spatial attentive single-image deraining with a high quality real
  rain dataset.
\newblock In {\em CVPR}, pages 12270--12279, 2019.

\bibitem{wang2021multi}
Yinglong Wang, Chao Ma, and Bing Zeng.
\newblock Multi-decoding deraining network and quasi-sparsity based training.
\newblock In {\em CVPR}, pages 13375--13384, 2021.

\bibitem{wang2004image}
Zhou Wang, Alan~C Bovik, Hamid~R Sheikh, and Eero~P Simoncelli.
\newblock Image quality assessment: from error visibility to structural
  similarity.
\newblock {\em IEEE TIP}, 13(4):600--612, 2004.

\bibitem{wang2022uformer}
Zhendong Wang, Xiaodong Cun, Jianmin Bao, Wengang Zhou, Jianzhuang Liu, and
  Houqiang Li.
\newblock Uformer: A general u-shaped transformer for image restoration.
\newblock In {\em CVPR}, pages 17683--17693, 2022.

\bibitem{wei2019semi}
Wei Wei, Deyu Meng, Qian Zhao, Zongben Xu, and Ying Wu.
\newblock Semi-supervised transfer learning for image rain removal.
\newblock In {\em CVPR}, pages 3877--3886, 2019.

\bibitem{xiao2022image}
Jie Xiao, Xueyang Fu, Aiping Liu, Feng Wu, and Zheng-Jun Zha.
\newblock Image de-raining transformer.
\newblock {\em IEEE TPAMI}, 2022.

\bibitem{2020learning}
Fuzhi Yang, Huan Yang, Jianlong Fu, Hongtao Lu, and Baining Guo.
\newblock Learning texture transformer network for image super-resolution.
\newblock In {\em CVPR}, 2020.

\bibitem{yang2017deep}
Wenhan Yang, Robby~T Tan, Jiashi Feng, Jiaying Liu, Zongming Guo, and Shuicheng
  Yan.
\newblock Deep joint rain detection and removal from a single image.
\newblock In {\em CVPR}, pages 1357--1366, 2017.

\bibitem{yang2020single}
Wenhan Yang, Robby~T Tan, Shiqi Wang, Yuming Fang, and Jiaying Liu.
\newblock Single image deraining: From model-based to data-driven and beyond.
\newblock {\em IEEE TPAMI}, 2020.

\bibitem{yasarla2020syn2real}
Rajeev Yasarla, Vishwanath~A Sindagi, and Vishal~M Patel.
\newblock Syn2real transfer learning for image deraining using gaussian
  processes.
\newblock In {\em CVPR}, pages 2726--2736, 2020.

\bibitem{ye2021closing}
Yuntong Ye, Yi Chang, Hanyu Zhou, and Luxin Yan.
\newblock Closing the loop: Joint rain generation and removal via disentangled
  image translation.
\newblock In {\em CVPR}, pages 2053--2062, 2021.

\bibitem{yi2021structure}
Qiaosi Yi, Juncheng Li, Qinyan Dai, Faming Fang, Guixu Zhang, and Tieyong Zeng.
\newblock Structure-preserving deraining with residue channel prior guidance.
\newblock In {\em ICCV}, pages 4238--4247, 2021.

\bibitem{yuan2021incorporating}
Kun Yuan, Shaopeng Guo, Ziwei Liu, Aojun Zhou, Fengwei Yu, and Wei Wu.
\newblock Incorporating convolution designs into visual transformers.
\newblock In {\em ICCV}, pages 579--588, 2021.

\bibitem{zamir2022restormer}
Syed~Waqas Zamir, Aditya Arora, Salman Khan, Munawar Hayat, Fahad~Shahbaz Khan,
  and Ming-Hsuan Yang.
\newblock Restormer: Efficient transformer for high-resolution image
  restoration.
\newblock In {\em CVPR}, pages 5728--5739, 2022.

\bibitem{zamir2021multi}
Syed~Waqas Zamir, Aditya Arora, Salman Khan, Munawar Hayat, Fahad~Shahbaz Khan,
  Ming-Hsuan Yang, and Ling Shao.
\newblock Multi-stage progressive image restoration.
\newblock In {\em CVPR}, pages 14821--14831, 2021.

\bibitem{zhang2017convolutional}
He Zhang and Vishal~M Patel.
\newblock Convolutional sparse and low-rank coding-based rain streak removal.
\newblock In {\em WACV}, pages 1259--1267, 2017.

\bibitem{zhang2018density}
He Zhang and Vishal~M Patel.
\newblock Density-aware single image de-raining using a multi-stream dense
  network.
\newblock In {\em CVPR}, pages 695--704, 2018.

\bibitem{zhang2019image}
He Zhang, Vishwanath Sindagi, and Vishal~M Patel.
\newblock Image de-raining using a conditional generative adversarial network.
\newblock {\em IEEE TCSVT}, 30(11):3943--3956, 2019.

\bibitem{zhang2023accurate}
Jiale Zhang, Yulun Zhang, Jinjin Gu, Yongbing Zhang, Linghe Kong, and Xin Yuan.
\newblock Accurate image restoration with attention retractable transformer.
\newblock {\em ICLR}, 2023.

\bibitem{zhao2019explicit}
Guangxiang Zhao, Junyang Lin, Zhiyuan Zhang, Xuancheng Ren, Qi Su, and Xu Sun.
\newblock Explicit sparse transformer: Concentrated attention through explicit
  selection.
\newblock {\em ICLR}, 2020.

\end{thebibliography}
}

\end{document}